\documentclass[10pt,journal,compsoc]{IEEEtran}
\usepackage{graphicx}
\usepackage{amsmath}
\usepackage{amssymb}
\usepackage{booktabs}
\usepackage{multirow}
\usepackage{color}
\usepackage[ruled,vlined]{algorithm2e}
\usepackage{amsmath}
\usepackage{algorithmic}
\usepackage[misc]{ifsym}
\usepackage{booktabs}
\usepackage{enumitem}

\ifCLASSOPTIONcompsoc
  \usepackage[nocompress]{cite}
\else
  \usepackage{cite}
\fi
\ifCLASSINFOpdf
\else
\fi
\begin{document}
%
\title{Diagnose Like a Radiologist: \\ Hybrid Neuro-Probabilistic Reasoning for Attribute-Based Medical Image Diagnosis}
%
%
%
%

\author{Gangming~Zhao,~\IEEEmembership{}
        Quanlong~Feng,~\IEEEmembership{}
        Chaoqi~Chen,~\IEEEmembership{}
        Zhen Zhou,~\IEEEmembership{}
        and~Yizhou~Yu,~\IEEEmembership{Fellow,~IEEE}
\IEEEcompsocitemizethanks{\IEEEcompsocthanksitem (Corresponding author: Yizhou Yu)
\IEEEcompsocthanksitem G. Zhao, C. Chen and Y. Yu are with Department of Computer Science, the University of Hong Kong, Pokfulam, Hong Kong. E-mail: yizhouy@acm.org
\IEEEcompsocthanksitem Q. Feng is with Department of Geographical Information Engineering, China Agricultural University, Beijing, China.
\IEEEcompsocthanksitem Z. Zhou is with the AI Lab, Deepwise Healthcare, Beijing 100080, China.

}
}

%
%

\markboth{}%
{Zhao \MakeLowercase{\textit{et al.}}: Neuro-Probabilistic Reasoning for Medical Image Diagnosis}
%



\IEEEtitleabstractindextext{%
\begin{abstract}
During clinical practice, radiologists often use attributes, e.g. morphological and appearance characteristics of a lesion, to aid disease diagnosis. Effectively modeling attributes as well as all relationships involving attributes could boost the generalization ability and verifiability of medical image diagnosis algorithms. In this paper, we introduce a hybrid neuro-probabilistic reasoning algorithm for verifiable attribute-based medical image diagnosis. There are two parallel branches in our hybrid algorithm, a Bayesian network branch performing probabilistic causal relationship reasoning and a graph convolutional network branch performing more generic relational modeling and reasoning using a feature representation. Tight coupling between these two branches is achieved via a cross-network attention mechanism and the fusion of their classification results. We have successfully applied our hybrid reasoning algorithm to two challenging medical image diagnosis tasks. On the LIDC-IDRI benchmark dataset for benign-malignant classification of pulmonary nodules in CT images, our method achieves a new state-of-the-art accuracy of 95.36\% and an AUC of 96.54\%. Our method also achieves a 3.24\% accuracy improvement on an in-house chest X-ray image dataset for tuberculosis diagnosis. Our ablation study indicates that our hybrid algorithm achieves a much better generalization performance than a pure neural network architecture under very limited training data.
\end{abstract}

\begin{IEEEkeywords}
Bayesian Networks, Deep Neural Networks, Medical Image Analysis, Neuro-Probabilistic Reasoning
\end{IEEEkeywords}}

\maketitle

\IEEEdisplaynontitleabstractindextext

%
\IEEEpeerreviewmaketitle

\ifCLASSOPTIONcompsoc

\IEEEraisesectionheading{\section{Introduction}\label{sec:introduction}}

\IEEEPARstart{D}{ue} to their rapid progress in the past ten years, deep neural networks have achieved tremendous success in boosting the performance of image recognition \cite{Ref:Krizhevsky2012,Ref:Russakovsky2015,he2016deep,Ref:He2016identity} and other visual computing tasks \cite{ren2015faster,long2015fully}. Such progress has also propelled forward many related research areas including medical image analysis. Nowadays, deep neural networks have been widely used for medical image analysis and diagnosis. They have demonstrated unprecedented accuracy on a variety of tasks, such as the detection of acute intracranial hemorrhage in head CT images~\cite{kuo2019expert}, breast cancer diagnosis using mammography~\cite{mckinney2020international}, the detection of diabetic retinopathy in retinal fundus photographs~\cite{gulshan2016development}, and the classification of skin cancers~\cite{esteva2017correction}.

Despite the high accuracy deep neural networks have achieved in medical image diagnosis, they still exhibit two major limitations, insufficient generalization ability and verifiability. It is well known that deep neural networks require large amounts of training data. When high quality training data is scarce, as is typically the case in the medical image domain, the performance of trained deep neural networks on novel testing data may deteriorate significantly. In addition, deep neural networks typically produce a final result without showing how and why the result can be reached. This is undesired because very often computer-generated diagnostic results only serve as decision support for human doctors, who need to verify their credibility before accepting them.

To improve verifiability, a medical image diagnosis algorithm needs to learn the knowledge reasoning process followed by radiologists during clinical practice. They start from visual evidences in medical images and reach diagnostic conclusions by referring to causal relationships between diseases and the visual evidences. For example, radiologists use intuitive morphological and appearance characteristics of a pulmonary nodule, such as lobulation, spiculation, and texture, as visual evidences to assess whether the nodule is benign or malignant~\cite{hussein2017risk} (Fig.~\ref{fig:att-lidc}). In this paper, such characteristics of a lesion or, in general, radiological abnormalities in a medical image are defined as the attributes of the lesion or image. Thus, equipping a machine learning model with the capability of simultaneously predicting attributes as well as underlying diseases improves the causality and verifiability of the model now that radiologists can verify the predicted attributes as well as the knowledge reasoning process that goes from attributes to diagnostic conclusions.

\begin{figure*}[t]
\begin{center}
\includegraphics[width=1\linewidth]{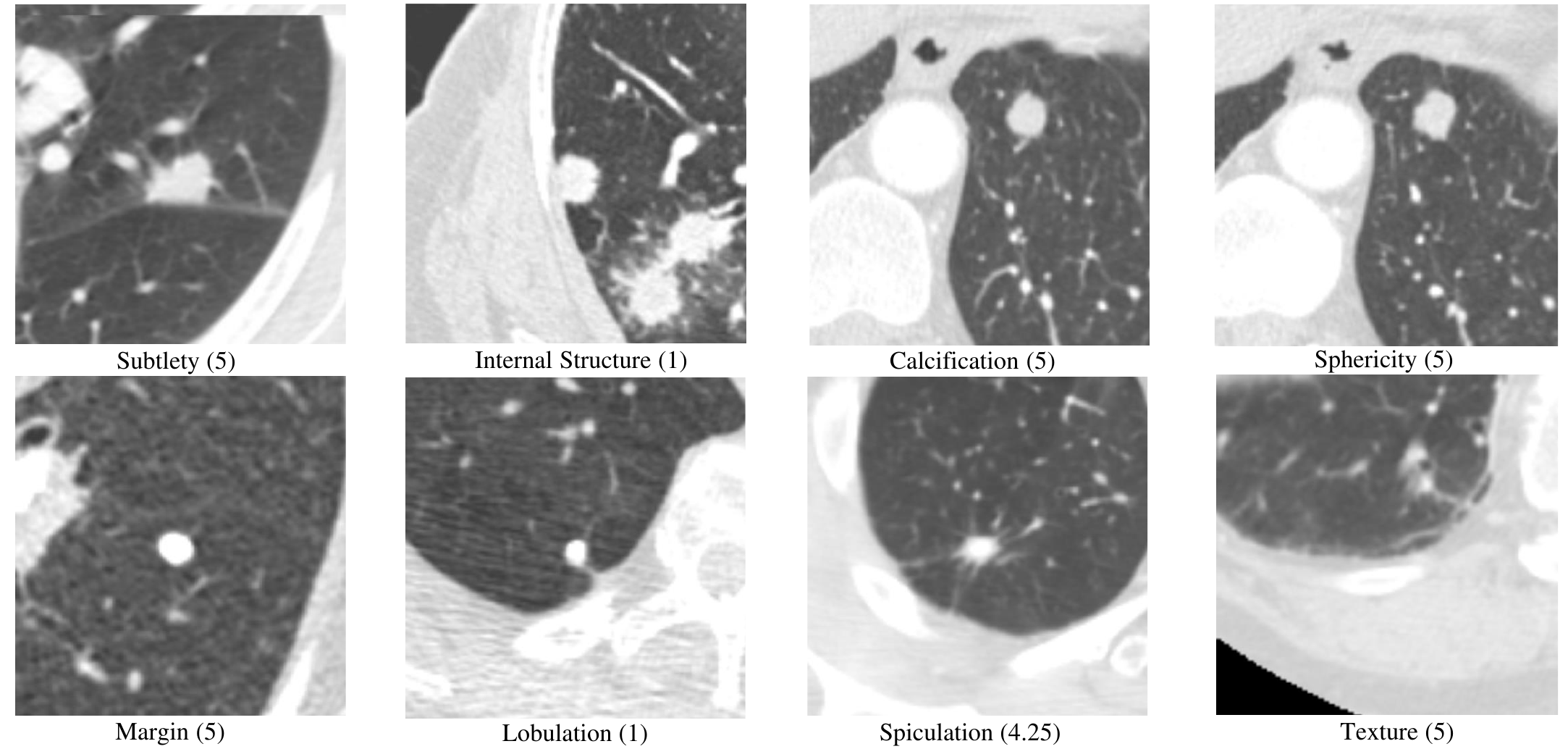}
\end{center}
\caption{Sample 2D slices of 3D pulmonary nodules from the LIDC-IDRI dataset, each with one of the following eight attributes: Subtlety, Internal Structure, Calcification, Sphericity, Margin, Lobulation, Spiculation, Texture. The annotated grade of each attribute is shown between the parentheses. For example, Subtlety (5) means its annotated grade is 5.}\label{fig:att-lidc}
\end{figure*}

In addition to causal relationships between diseases and attributes, there also exist relationships, such as co-occurrence, among the attributes themselves. Effectively modeling and exploiting the hidden relationships among these attributes as well as the relationships between attributes and the underlying diseases can boost the accuracy of attribute and disease prediction. Nonetheless, there have been few methods in medical image diagnosis exploiting such relationships.

To address generalization ability, we need to look at the strengths and limitations of both neural and probabilistic learning algorithms for relational modeling and reasoning. Examples of neural algorithms include graph neural networks~\cite{kipf2016semi} and relation networks~\cite{santoro2017simple} while examples of probabilistic learning algorithms include Bayesian networks~\cite{pearl2014probabilistic} and factor graphs~\cite{frey1997factor}.  Neural algorithms are good at learning from large amounts of empirical data. The trained models can interpolate and approximate known information very well to reach high accuracy, but have relatively poor generalization ability, causality and explainability. On the other hand, probabilistic learning algorithms, which can be considered as a type of symbolic AI methods, can learn from smaller datasets. The trained models have better generalization ability, causality and explainability, but lower accuracy when facing real-world empirical data.

In this paper, to exploit the complementary strengths of neural and probabilistic learning algorithms as well as overcome the limitations of both ends, we introduce a hybrid neuro-probabilistic reasoning algorithm for verifiable attribute-based medical image diagnosis. Specifically, the proposed hybrid algorithm consists of multiple stages. It first employs a deep neural network backbone to extract features from an input image. Such features build the foundation of reasoning that happens in later stages. Next, a Bayesian network (BN) branch and a graph convolutional network (GCN) branch process the features extracted by the backbone and perform relational reasoning in parallel. The BN branch first converts features into attributes as evidences, then performs probabilistic reasoning using the causal relationships between the disease and attributes, and finally produces their marginal posterior distributions. On the other hand, the GCN branch performs feature-based relational modeling. It represents attributes and generic relationships among them using a graph. The GCN branch exploits the relationships among the graph nodes to enhance the feature of every attribute. More accurate diagnoses and attribute classifications can be achieved using the enhanced features from the GCN. Although the BN and GCN belong to two parallel branches, they are not completely independent. We tightly couple the BN and GCN branches so that the BN is utilized to improve the causality and generalization ability of the GCN while the accuracy of the GCN is retained. Such tight coupling is achieved via a cross-network attention mechanism and the fusion of their classification results. By fusing the results from the two branches, causality enforced by the BN branch may be partially compromised. Therefore, a second Bayesian network module is appended at the end of the entire network to reinforce causality.

The entire network except the Bayesian network modules can be trained using gradient back-propagation as the inference phase of a Bayesian network is differentiable and gradients can be propagated through the two Bayesian networks in our algorithm. However, we notice that the compatibility between the structure of the Bayesian networks and the rest of our hybrid network significantly affects the performance of the entire network. Thus, once we learn the initial structure of the Bayesian networks using ground-truth annotations, the training procedure proceeds by alternating between two phases. The first phase updates the CNN and GCN weights using stochastic gradient descent while the second phase updates the structures and parameters of the two Bayesian networks using the attribute classification results most recently predicted by their preceding stages in our hybrid network as training labels.

Note that verifiability is more stringent than explainability. For computer-generated diagnoses to be verifiable, the internal decision process of an algorithm not only should be explainable, but also needs to comply with the domain knowledge of human doctors.
As mentioned earlier, in medical image diagnosis, radiologists start from visual evidences in medical images and reach diagnostic conclusions by referring to causal relationships between diseases and the visual evidences.
Thus the algorithm needs to collect the same types of visual evidences as a doctor would do, and also learn the causal relationships between evidences and diseases as a doctor would do.
The proposed algorithm in this paper models visual evidences as attributes and collects such visual evidences through attribute prediction.
It further models causal and non-causal relationships using BN and GCN, which quantitatively encode such medical domain knowledge as ``the presence of attribute A in this CT scan increases/decreases the likelihood that the patient has disease D" or ``the occurrence of attribute A in this CT scan increases/decreases the chance that attribute B also occurs in the same scan".
Every node in these networks (BN and GCN) represents a high-level concept (a disease or an attribute), and a disease diagnosis exploits the relationships between the disease and a set of other high-level concepts.
In comparison to previous medical image diagnosis methods, radiologists may find the knowledge reasoning process of the proposed algorithm more consistent with their own practice and, consequently, the disease diagnoses made by our method more credible.

We have successfully applied our hybrid reasoning algorithm to two challenging medical image diagnosis tasks. The first task is benign-malignant classification of pulmonary nodules in chest computed tomography (CT) images in the LIDC-IDRI benchmark dataset. As mentioned earlier, for pulmonary nodule classification, morphological and appearance attributes have been commonly used to assist clinical diagnosis~\cite{hussein2017risk}. The second task is tuberculosis (TB) diagnosis using chest X-ray images. For this task, we adopt an in-house dataset carefully annotated by senior medical experts. Every image in this TB dataset is not only annotated with a disease diagnosis, but also multiple types of radiological abnormalities, such as `Pulmonary Consolidation', `Pulmonary Cavitation', `Diffuse Nodules', and `Fibrotic Appearance', potentially caused by tuberculosis.
According to the experimental results of pulmonary nodule classification on the LIDC-IDRI benchmark, our method achieves a new state-of-the-art accuracy of 95.36\% and an AUC of 96.54\%. On our in-house dataset for tuberculosis classification, our algorithm also achieves a 4.98\% improvement in accuracy in comparison to previously best-performing algorithm for attribute-based medical image classification.

Our main contributions in this paper can be summarized as follows.
\begin{itemize}
\item We introduce a hybrid neuro-probabilistic reasoning algorithm for attribute-based medical image diagnosis. To support causal and verifiable relational modeling and reasoning, this algorithm tightly couples Bayesian networks and a graph convolutional network. Such coupling is achieved via a cross-network attention mechanism and a classification result fusion scheme.
\item We devise an effective training procedure for the proposed hybrid network. It alternates between two phases. The first phase updates the CNN backbone and GCN weights while the second phase updates the structures and parameters of the Bayesian networks.
\item We have successfully applied the proposed hybrid reasoning algorithm to two representative medical image diagnosis tasks, benign-malignant classification of pulmonary nodules in chest CT images and tuberculosis diagnosis using chest X-ray images. The proposed algorithm achieves state-of-the-art performance on the LIDC-IDRI benchmark dataset for the first task and an in-house dataset for the second task.
\end{itemize}

\section{Related Work}
\subsection{Attribute Learning}
Attributes, such as texture, color, and shape, are of great importance to describe objects. Attribute learning has been studied in computer vision for many years~\cite{ferrari2008learning,kumar2009attribute,akata2013label,lampert2013attribute,liang2017incomplete,liang2018unifying,min2019multi}. Ferrari et al.~\cite{ferrari2008learning} proposed to use low-level semantic features for attribute representation and they presented a probabilistic generative model for visual attributes, together with an image likelihood learning algorithm. Human faces have many attributes, and remain a challenge for attribute learning. Kumar et al.~\cite{kumar2009attribute} trained binary classifiers to recognize the presence or absence of describable aspects of facial visual appearance using traditional hand-crafted features. Liu et al.~\cite{liu2015deep} proposed a CNN framework for face localization and attribute prediction, respectively.
Attributes have also been exploited in tasks such as zero-shot learning~\cite{lampert2013attribute,jiang2017learning}.
Effectively modeling the hidden relationships among attributes is useful for improving the accuracy of attribute prediction and causal association. Nonetheless, most of the early works in attribute learning did not model relationships among attributes and explore such relationships for attribute reasoning. 
The development of graph neural networks (GNN)~\cite{kipf2016semi} made it possible to learn relationships among attributes. For example, Meng et al.~\cite{meng2018efficient} used message passing to perform end-to-end learning of image representations, their relationships as well as the interplay among different attributes. They observed that relative attribute learning naturally benefits from exploiting the graph of dependencies among different image attributes. In this paper, we not only utilize a graph neural network to model the correlations among attributes, but also embed Bayesian networks into the whole framework to better model causality.

\begin{figure*}[t]
\begin{center}
\includegraphics[width=1\linewidth]{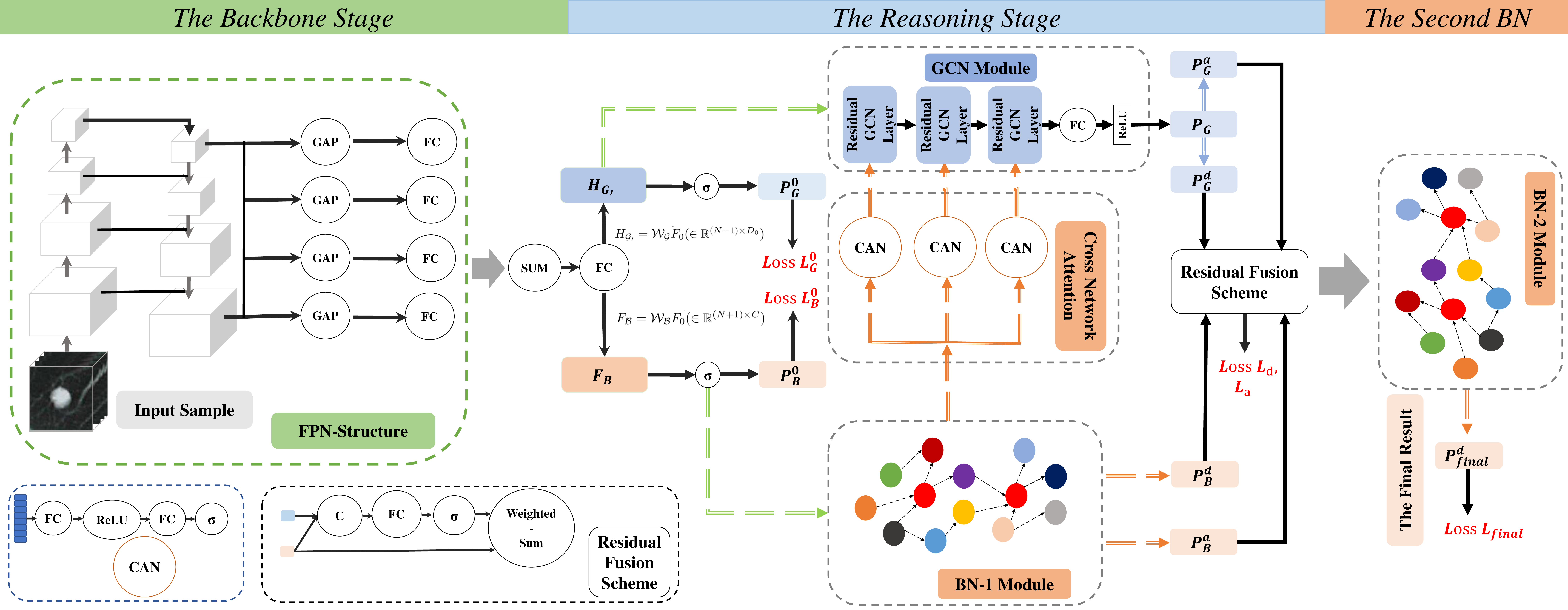}
\end{center}
\caption{Our proposed framework consists of three stages: 1) Deep feature extraction using an CNN-FPN backbone; 2) hybrid attribute relational modeling and reasoning using a pair of coupled Bayesian network and deep graph network; 3) causality reinforcement using a second Bayesian network. Note that BN-1 and BN-2 are distinct in both parameters and structures.}\label{fig:framework}
\end{figure*}

\subsection{Bayesian Networks}
Bayesian networks (BN), introduced by Judea Pearl~\cite{pearl2011bayesian}, represent a natural approach to model causality and perform logical reasoning. Bayesian network learning includes two phases, structure learning and parameter learning. The most intuitive method for structure learning is that of `search and score,' where one searches the space of directed acyclic graphs (DAGs) using dynamic programming and identifies the one that minimizes the objective function~\cite{wit2012all}. For parameter learning, the most frequently used method is maximum likelihood estimation (MLE)~\cite{pearl2011bayesian}. However, when given a dataset, BN cannot learn a feature representation, which limits its further development. Recently, researchers have started to focus on the integration between BN and deep learning. For example, Rohekar et al.~\cite{rohekar2018constructing} proposed to utilize BN models for learning better deep neural networks. Meanwhile, a few improved versions of BN have been proposed for new applications in computer vision. For example, Barik~\cite{barik2019learning} improved BN using low-rank conditional probability tables. Elidan et al.~\cite{elidan2010copula} used the Copula Bayesian Network model for representing multivariate continuous distributions. The method in this paper differs from the above work in that it exploits the representation power of deep neural networks and the causality modeling capability of BN by embedding them into a unified framework.

\subsection{Bayesian Neural Networks}
In addition to poor interpretability, deep neural networks also suffer from overfitting on limited labeled data and the incapability of uncertainty analysis. To tackle these issues, researchers combined Bayesian methods with deep learning to construct Bayesian neural networks (BNN) for a probabilistic representation of uncertainty of the latter~\cite{kingma2015variational,gal2016dropout,shi2018bayesian}. Specifically, a BNN estimates posterior distributions for the weights in a neural network. There exist a few estimation methods, such as variational inference~\cite{kingma2015variational} and dropout variational inference~\cite{gal2016dropout}.
However, the proposed method in this paper differs significantly from Bayesian neural networks. Unlike BNNs, our proposed method aims to improve the causality and interpretability of deep learning via the fusion of deep learning and traditional graph-based probabilistic reasoning. It cannot perform uncertainty analysis. Nonetheless, due to the few trainable parameters in Bayesian networks, the proposed method can clearly mitigate overfitting on small training datasets, as demonstrated in one of our ablation studies.

\subsection{Neuro-symbolic Learning}
The integration of connectionist (neural networks) and symbolic AI has long been a key issue in machine learning~\cite{shavlik1991symbolic,shavlik1992framework,yi2018neural,yin2018structvae,li2020closed}. Early works~\cite{shavlik1991symbolic,shavlik1992framework} primarily focused on the injection of symbolic information as prior knowledge into a neural network. More recently, with the wide spread of deep learning, researchers~\cite{yi2018neural,yin2018structvae,li2020closed} started to combine the merits of deep feature representation with symbolic reasoning to build strong, reliable and explainable neural-symbolic AI systems. Relevant studies have been conducted on visual question answering (VQA)~\cite{yi2018neural}, semantic parsing~\cite{yin2018structvae} and handwritten formula recognition~\cite{li2020closed}.
One of the key issues in neural-symbolic learning is that most symbolic approaches are not differentiable, making end-to-end training difficult. In this study, we partially solved this issue by performing gradient back-propagation through both Bayesian networks, which serve as the symbolic reasoning module in our hybrid network, for a better feature representation and a more efficient training procedure.

\section{Algorithm}

\subsection{Overview}
The proposed hybrid neuro-probabilistic reasoning algorithm takes an image patch as the input and outputs a classification result, as shown in Fig.~\ref{fig:framework}. The proposed algorithm consists of the following three stages.

{\flushleft \bf The Backbone Stage:} a backbone based on ResNet~\cite{he2016deep} or EfficientNet~\cite{tan2019efficientnet} is utilized for deep feature representation, which builds the foundation for reasoning in later stages. Moreover, a feature pyramid network (FPN)~\cite{lin2017feature} is integrated into the backbone for multi-scale feature aggregation (Fig.~\ref{fig:framework}).
A distinct global average pooling (GAP) operator is applied to the feature map at each scale of the FPN to reduce the number of parameters. And for reducing information loss in low-level feature maps, we ensure that the dimension of the resulting feature from GAP is kept the same at different scales of the FPN. For instance, when the size of a feature map in the FPN is $64 \times 64 \times 256$, we first reshape it to $32 \times 16 \times 2048$ by concatenating the third dimensions of every $2\times 4$ neighbors in the first two dimensions together, then use GAP to produce a 2048-dimensional feature vector. All GAP-reduced features go through parallel FC layers before they are fused via a sum operator and a subsequent FC layer to obtain the final feature, $F_{0}$, at the backbone stage.

{\flushleft \bf The Reasoning Stage:} both a Bayesian network and a graph convolutional network are adopted for relational reasoning in parallel. The BN branch first converts deep features into attributes through attribute classification, then performs probabilistic causal reasoning using these attributes, and finally outputs marginal posterior distributions. The GCN branch first transforms the deep features from the backbone into a feature representation for individual attributes and represents the relationships among the attributes using an undirected graph, then enhances the feature of each attribute via graph convolutions. The BN and GCN are tightly coupled together through a cross-network attention mechanism and the fusion of their classification results, which will be described in Section~\ref{sec:integration}.

{\flushleft \bf The Second BN Module:} to reinforce causality of the proposed hybrid algorithm, a second BN is appended at the end of the entire network.

\subsection{Integration between BN and Backbone}
A Bayesian network is incorporated in the reasoning stage to model probabilistic dependencies among various attributes for disease diagnosis. Specifically, a Bayesian network, $\mathcal{B} = <\mathcal{V_{B}}, \mathcal{E_B}, \Theta> $, is a directed acyclic graph (DAG) $<\mathcal{V_{B}}, \mathcal{E_B}> $ with a conditional probability table (CPT) for each node, and $\Theta$ represents all parameters (CPTs), which encodes the joint probability distribution of the BN. Each node $v_i \in \mathcal{V_B}$ stands for a random variable, and a directed edge $e \in \mathcal{E_B}$ between two nodes $(v_i, v_j)$ indicates $v_i$ probabilistically depends on $v_j$.
The training of a Bayesian network undergoes two phases, structure learning and parameter learning~\cite{pearl2011bayesian,pearl2014probabilistic,eaton2012bayesian}. We adopt dynamic programming with the Bayesian information criterion (BIC)~\cite{wit2012all} for structure learning to determine the topology of the DAG and maximum likelihood estimation for parameter learning to determine $\Theta$. Our BN module first learns its structure and then updates all parameters, i.e. the conditional probability tables at all nodes in the network.
In this paper, each attribute has a corresponding node in the Bayesian network. Without loss of generality, assume there is only one disease we wish to diagnose, and that disease is an extra node in the network.

To integrate the Bayesian network with our backbone, the final feature from the backbone, $F_{0}$, is fed to another FC-layer with weight matrix $\mathcal{W}_{\mathcal{B}}$ to obtain a feature map $F_{\mathcal{B}}$. Thus, $F_{\mathcal{B}} = \mathcal{W}_{\mathcal{B}} F_{0} (\in \mathbb{R}^{(N+1) \times C})$, where $N$ is the number of attribute categories and $C$ is the number of grades for each attribute or disease. Each attribute must have at least two grades to indicate whether that attribute is present or not. In practice, there is often a need to have more than two grades for each attribute or disease to indicate intermediate levels of certainty or severity. Note that different grades of the same attribute are mutually exclusive while different attributes are not because more than one attribute could be present at the same time.
Thus, $F_{\mathcal{B}}$ can be interpreted as the concatenation of $N+1$ $C$-dimensional vectors, one for each node in the Bayesian network.
We further define $P^0_{B} = \sigma(F_{\mathcal{B}}) (\in [0, 1]^{(N+1) \times C})$, where the softmax operator $\sigma$ is applied to each $C$-dimensional vector in $F_{B}$, and the result represents a discrete distribution over the grades of an attribute or disease. This distribution becomes the input evidence at the corresponding node in the Bayesian network.




Once we perform inference in the Bayesian network using the conditional probability tables as well as the input evidences at all nodes, we obtain marginal posterior distributions at all nodes, including the extra node for disease diagnosis, as the output of the Bayesian network.
Let the nodes in the Bayesian network be $v_{i}, i \in \{0,1,...,n\}$. Without loss of generality, according to \cite{pearl2011bayesian}, the marginal posterior distribution $P_B(v_0)$ at node $v_0$ is formulated as
\begin{equation}
P_B(v_{0}) = \int...\int_{V} P(v_{0}, v_{1},..., v_{n}) \mathrm{d}v_{1}...\mathrm{d}v_{n},
\label{eq:bn-att-multi}
\end{equation}
where
\begin{equation}
P(v_{0}, v_{1},..., v_{n}) =  \Pi_{i=0}^{n} P(v_{i} | Parents(v_{i})),
\label{eq:bn-att-joint}
\end{equation}
where $Parents(v_{i})$ is NULL if $v_{i}$ does not have any parent nodes. The equation in (\ref{eq:bn-att-joint}) is derived using the local Markov property of Bayesian networks. In practice, to evaluate (\ref{eq:bn-att-multi}), we use the belief propagation algorithm in \cite{pearl1982reverend}.

\subsection{GCN Module}
In addition to the Bayesian network, we also adopt a graph convolutional network in the reasoning stage to perform feature-based relational modeling and reasoning among attributes to facilitate medical image diagnosis. Specifically, a GCN is based on an undirected graph $\mathcal{G}=(\mathcal{V_{G}}, \mathcal{E_G})$, where $\mathcal{V_{G}}$ and $\mathcal{E_G}$ are the set of nodes and edges respectively. Here, there is a node corresponding to each attribute or disease, and the node holds a feature vector associated with that attribute or disease. Again, assume there is only one disease we wish to diagnose, and the total number of nodes is $N+1$.
If vertices $v_{i}$ and $v_{j}$ are connected in the graph, there might exist certain type of relations between them.
We associate a learnable weight with every edge in the graph to indicate the strength of the connection.
The GCN in our algorithm consists of $L$ layers, all of which share the same number of nodes. But the edge weights in these layers may be different.

A general graph convolution operation $\mathcal{F}$ at the $l$-th layer can be formulated as follows
\begin{equation}
\mathcal{F}(\mathcal{G}_{l}, \mathcal{W}_{l}) = \mbox{Update}(\mbox{Aggregate}(\mathcal{G}_{l}, \mathcal{W}^{agg}_{l}), \mathcal{W}^{update}_{l}),
\label{BaseGraph}
\end{equation}
where $\mathcal{G}_{l} = (\mathcal{V_{G}}_{l}, \mathcal{E}_{l})$ is the graph at the $l$-th layer; $\mathcal{W}^{agg}_{l}$ and
$\mathcal{W}^{update}_{l}$ are the learnable weights of the aggregation and update functions of the $l$-th layer, respectively.
We use a max-pooling node feature aggregator to pool the pairwise differences of features at node $v_{i}$ and all of its neighbors. But the feature difference between $v_{i}$ and one of its neighbors is modulated by the edge weight between them before the pooling operation. The node feature updater is a multi-layer perceptron (MLP) with batch normalization and ReLU as the activation function. The above MLP concatenates the original node feature with the aggregated feature as its input.

We adopt the residual graph convolution introduced in \cite{li2019deepgcns}.
Let $H_{\mathcal{G}_{l}} \in \mathcal{R}^{N \times D_{l}}$ be the feature map holding the features at all nodes of $\mathcal{G}_{l}$, where $N$ is the number of attributes and $D_{l}$ is the feature dimension in the $l$-th layer.
The residual graph convolution between the $l$-th and ($l+1$)-th layers of the GCN is formulated as follows,
\begin{equation}
H_{\mathcal{G}_{l+1}} = \mathcal{F}(\mathcal{G}_{l}, \mathcal{W}_{l}) + H_{\mathcal{G}_{l}}.
\label{ResGraph}
\end{equation}

To integrate the above GCN with our backbone, we need to obtain $N+1$ features from the final feature in the backbone, one for each node in the GCN. For this purpose, $F_{0}$ is fed to another FC-layer with weight matrix $\mathcal{W}_{\mathcal{G}}$ to obtain feature map $H_{\mathcal{G_0}}$. Thus, $H_{\mathcal{G_0}} = \mathcal{W}_{\mathcal{G}} F_{0} (\in \mathbb{R}^{(N+1) \times D_0})$, where $N$ is the number of attribute categories and $D_0$ is the input feature dimension for every node in the GCN. $H_{\mathcal{G_0}}$ thus can be interpreted as the concatenation of $N+1$ $D_0$-dimensional feature vectors, one for each node in the first layer of the GCN as input.


The final feature map of the GCN is $H_{\mathcal{G}_{L}} \in \mathbb{R}^{(N+1) \times D_{L}}$. We further use a FC-layer followed by a softmax operator to obtain attribute and disease classification results in the following equation,
\begin{equation}
P_G = \sigma(\mathcal{W}_{P}H_{\mathcal{G}_{L}})
\label{eq:pgupdate}
\end{equation}
where $P_G \in [0,1]^{(N+1) \times C}$ denotes the attribute and disease predictions.

\subsection{Coupling between BN and GCN}\label{sec:integration}
Although both the BN and GCN are utilized for relational reasoning, they are not completely independent. In this work, we tightly couple BN and GCN via a cross-network attention mechanism and the fusion of classification results. As for the former, BN provides attention for the node features in each layer of the GCN. Specifically, the marginal posterior distributions at all nodes of the BN are taken as the input of $L$ node attention modules, one for each layer of the GCN to figure out the relative importance of attributes in that layer.
Node attention values prescribed by the $l$-th attention module are defined as follows,
\begin{equation}
M_{attn}^{l} = \sigma( \mathcal{W}_{l_{1}}\mbox{ReLU}(\mathcal{W}_{l_{0}}P_B(\mathcal{V})) ), l \in {1,2,...,L}
\label{sattn}
\end{equation}
where $M_{attn}^{l} \in \mathcal{R}^{N+1}$ is the node attention vector for the $l$-th layer of the GCN, $\mathcal{V}$ includes all attribute and disease nodes of the BN, and $P_B(\mathcal{V}) \in \mathcal{R}^{(N+1) \times C}$ represents the concatenated marginal distributions at all nodes of the BN.
Note that there is one attention value in $M_{attn}^{l}$ for each node in the $l$-th layer of the GCN. We name $M_{attn}^{l}$ the spatial attention map for GCN nodes since nodes are spatial entities of a graph, similar to pixels in a CNN.


We also extend squeeze-and-excitation based channel-wise attention in SENet~\cite{hu2018squeeze} to graph convolutional networks as follows  to further enhance the features of every layer in the GCN,
\begin{equation}
\begin{aligned}
C_{attn}^{l} = &  \sigma( \mathcal{W}_{l_{ex}}\mbox{ReLU}(\mathcal{W}_{l_{sq}}GAP(H_{\mathcal{G}_{l}})) )
\label{cattn}
\end{aligned}
\end{equation}
where ${C_{attn}^{l}} \in \mathcal{R}^{D_l}, l \in {1,2,...,L}$ is the channel-wise attention vector for the $l$-th layer of the GCN, $\mathcal{W}_{l_{sq}}$ and $\mathcal{W}_{l_{ex}}$ are the weights for squeezing and excitation, respectively.

Once both spatial and channel-wise attention vectors have been calculated, firstly the same channel of all features in $\mathcal{G}_{l}$ is multiplied by its corresponding channel-wise attention value, then the entire feature at each node in $\mathcal{G}_{l}$ is multiplied by its corresponding spatial attention value. Such spatially and channel-wise modulated features at the $l$-th layer of the GCN serve as the input to the next graph convolution to produce the feature map $H_{\mathcal{G}_{l+1}}$ at the $(l+1)$-th layer.

As for the fusion of the disease classification results from BN and GCN, a residual fusion scheme is formulated as follows,
\begin{equation}
P^{d}_{fusion} = w_{B} \cdot P^d_B + (1-w_{B}) \cdot \sigma(W_{0}\mbox{Concat}(P^d_G, P^d_B)).
\label{eq:fusedis}
\end{equation}
For attribute classification results, the same residual fusion scheme is adopted as follows,
\begin{equation}
P^{a}_{fusion} = w'_{B} \cdot P^a_B + (1-w'_{B}) \cdot \sigma(W'_{0}\mbox{Concat}(P^a_G, P^a_B)).
\label{eq:fuseattr}
\end{equation}
In these equations, $P^d_B$ and $P^a_B$ are the marginal posterior distributions at the disease and attribute nodes in the BN, $P^d_G$ and $P^a_G$ are the probabilistic disease and attribute classification results from the GCN, $w_{B}$ and $w'_{B}$ are two learnable trade-off coefficients, $W_0$ and $W'_0$ are the weight matrices of two fully connected layers, and Concat() is the concatenation operator.

\subsection{The Second BN Module}\label{sec:BN2}
It should be noted that by fusing the results of both BN and GCN branches, causality enforced by the BN branch may be compromised to some extent. To tackle this issue, we append a second Bayesian network at the end of the entire network to reinforce causality. The fused attribute and disease soft classification results in (\ref{eq:fuseattr}) and (\ref{eq:fusedis}) are used as the input evidences at the nodes in this second Bayesian network. The marginal posterior distribution at the disease node, denoted as $P^d_{final}$, can also be defined using equations similar to those in (\ref{eq:bn-att-multi}) and (\ref{eq:bn-att-joint}). In practice, it is also evaluated using the belief propagation algorithm. This marginal posterior distribution at the disease node of the second BN becomes the final disease prediction of the entire network. Note that BN-1 and BN-2 are distinct in both parameters and structures.

\subsection{Training Scheme}\label{sec:trainhrbrid}
Our entire network except the BN modules can be trained using stochastic gradient descent. Meanwhile, we observe that the inference phase of a BN is differentiable with respect to its inputs and parameters, but not its structure. That implies gradients can be backpropagated through a BN from its outputs (i.e. marginal posterior distributions) to its inputs (i.e. evidences). Parameter learning in a BN is tightly coupled with structure learning, and is driven by global statistics of the training set. Performing parameter learning separately from structure learning using stochastic gradient descent is unlikely to be very effective. Moreover, we discovered that the compatibility between the Bayesian network structures and the neural parts of our hybrid network significantly affects the overall performance.
Therefore, we propose the following two alternating phases to train the entire hybrid network.
\begin{itemize}
  \item The first phase updates the weights of the CNN backbone and the GCN branch using gradient backpropagation by fixing the structures and parameters of the two BNs. Gradients are simply backpropagated through these two BNs.
  \item  The second phase updates the structures and conditional probability tables of the BN modules using the attribute and disease classification results most recently predicted by their preceding stages in the hybrid network as training labels. As for structure learning, dynamic programming is used with the same score function as in \cite{wit2012all}. Once the structure of BN modules are determined, their conditional probability tables are updated immediately using maximum likelihood estimation.
\end{itemize}
It should be noted that we actually set a maximum value on the number of times the structures and parameters of the two BNs are updated (20 times in practice). Once the two phases have been alternated for such a number of times, we fix the structures and parameters of the two BNs and only update the weights in the neural parts of our hybrid network. In this way, we can guarantee the convergence of our training process.

The ground-truth annotations of the training samples are only used to train the initial structure and parameters of the BN modules. In subsequent iterations, for the first BN, the ground-truth attribute and disease annotations of a training sample are replaced with its attribute and disease classification results most recently produced by the backbone, i.e. $P^0_B$; for the second BN, they are replaced with its most recent fused attribute and disease classification results, $P^d_{fusion}$ and $P^a_{fusion}$ in (\ref{eq:fusedis}) and (\ref{eq:fuseattr}). The effectiveness of this training strategy has been empirically verified. Potential reasons for the effectiveness are twofold: first, attribute classification results produced at intermediate stages of our hybrid network are soft labels, which are actually distributions, and have more complete information than the ground-truth attribute labels represented as one-hot vectors; second, using soft attribute labels generated within the hybrid network makes the trained Bayesian network structures more compatible with the neural parts of the hybrid network.

We rely on the belief propagation algorithm to perform inference in a Bayesian network and compute the marginal posterior distributions for its variables. Since we train the weights in the neural networks by propagating gradients through the Bayesian networks, we need to obtain the gradient of the belief propagation algorithm. A method for computing this gradient is described in the Appendix.


\subsubsection{Training Loss}
During the first phase of the above alternating training procedure, we adopt the deep supervision strategy by distributing multiple supervision signals to various stages of our hybrid network. Most of the time, each supervision signal is primarily responsible for training a subset of the modules in the network. The overall loss function during the first phase of training is a weighted combination of five loss functions,
\begin{equation}\label{eq:loss-all}
\begin{aligned}
\mathcal{L}_{all} = w_{1}\mathcal{L}^0_{G} + &w_{2}\mathcal{L}^0_{B} +  w_{3}\mathcal{L}_{a} + w_{4}\mathcal{L}_{d} + w_{5}\mathcal{L}_{final}
\end{aligned}
\end{equation}
where $\sum^5_{i=1}w_{i} = 1$. In our experiments, we always set $w_{i}=0.2, i \in {1,2,..,5}$.

The first two supervision signals corresponding to losses $\mathcal{L}^0_{B}$ and $\mathcal{L}^0_{G}$ are respectively located at the beginning of the BN and GCN branches. Two classifiers are involved. One of them is simply $P^0_{B} = \sigma(F_{\mathcal{B}}) (\in [0, 1]^{(N+1) \times C})$, the input evidences at the nodes of the first Bayesian network. The cross-entropy loss for this classifier is written as follows,
\begin{equation}
\begin{aligned}
\mathcal{L}^0_{B} = & -\sum_{i=1}^{N+1}\sum_{j=1}^{C} (y_{ij}\log(p^{B,0}_{ij}) + (1 - y_{ij})\log(1 - p^{B,0}_{ij}))
\label{eq:loss-b}
\end{aligned}
\end{equation}
The second classifier is defined as $P^0_{G}=\sigma(\mathcal{W}_{\mathcal{T}}H_{\mathcal{G_0}}) (\in [0, 1]^{(N+1) \times C})$, which takes $H_{\mathcal{G_0}}$ as the input and performs both attribute and disease classification. The cross-entropy loss for this classifier is written as follows,
\begin{equation}
\begin{aligned}
\mathcal{L}^0_{G} = & -\sum_{i=1}^{N+1}\sum_{j=1}^{C} (y_{ij}\log(p^{G,0}_{ij}) + (1 - y_{ij})\log(1 - p^{G,0}_{ij}))
\label{eq:loss-g}
\end{aligned}
\end{equation}
In the above two loss functions, $y_{ij}$'s denote one-hot encoding of the ground-truth attribute or disease grades of training samples, $p^{B,0}_{ij}$ and $p^{G,0}_{ij}$ are components of $P^0_{B}$ and $P^0_{G}$ respectively. These two classification losses disentangle attributes from the deep features of the backbone under the supervision of ground-truth labels. These two supervision signals are primarily responsible for training the backbone as well as the connection layers between the backbone and the BN or GCN.

The second group of two supervision signals are applied to the fused results of the BN and GCN in (\ref{eq:fusedis}) and (\ref{eq:fuseattr}).
Two cross-entropy losses $\mathcal{L}_{a}$ and $\mathcal{L}_{d}$ are imposed again on the classification results of attributes and diseases, respectively.
The first loss is imposed on the fused disease classification result, $P^{d}_{fusion}$ in (\ref{eq:fusedis}), and is formulated as follows,
\begin{equation}
\begin{aligned}
\mathcal{L}_{d} = & -\sum_{j=1}^{C} (y^{d}_{j}\log(p^{d}_{j}) + (1 - y^{d}_{j})\log(1 - p^{d}_{j})),
\label{eq:loss-dis}
\end{aligned}
\end{equation}
where $y^{d}$ denotes one-hot encoding of the ground-truth disease grades of training samples, and $p^{d}_{j}$ represents a component of $P^{d}_{fusion}$.
The second loss $\mathcal{L}_{a}$ is imposed on the fused attribute classification results, $P^{a}_{fusion}$ in (\ref{eq:fuseattr}), and is formulated as follows,
\begin{equation}
\begin{aligned}
\mathcal{L}_{a} = & -\sum_{i=1}^{N}\sum_{j=1}^{C} (y^{a}_{ij}\log(p^{a}_{ij}) + (1 - y^{a}_{ij})\log(1 - p^{a}_{ij})),
\label{eq:loss-dis2}
\end{aligned}
\end{equation}
where $y^{a}$ denotes one-hot encoding of the ground-truth attribute grades of training samples, and $p^{a}_{ij}$ represents a component of $P^{a}_{fusion}$.
Since we do not update parameters in the BNs during the first phase of the training scheme, these two supervision signals are primarily responsible for training the backbone and the GCN.

The last supervision signal is located at the end of the entire hybrid network, and is responsible for training all modules in the entire network.
A cross-entropy loss is imposed on the final disease classification result, $P^d_{final}$, defined in Section~\ref{sec:BN2} and produced by the second Bayesian network. It is written as follows,
\begin{equation}
\begin{aligned}
\mathcal{L}_{final} = & -\sum_{j=1}^{C} (y^{d}_{j}\log(p^{final}_{j}) + (1 - y^{d}_{j})\log(1 - p^{final}_{j})),
\label{eq:loss-last-dis}
\end{aligned}
\end{equation}
where $p^{final}_{j}$ represents a component of $P^d_{final}$.


\begin{table*}[t]\normalsize
	\caption{Performance Comparison of Lung Nodule Classification Models on the LIDC-IDRI Dataset. `B' And `M' are the Number of Benign and Malignant Lung Nodules Used for Model Training.}
\scalebox{0.75}{%
	\centering
\begin{tabular}{c|c|c|c|c|c|c|c|c|c}
\toprule[2pt]
  & \multirow{2}{*}{Methods} &  \multicolumn{2}{|c|}{Number} & \multicolumn{6}{|c}{Results(\%) (mean$\pm$standard deviation)} \\
 \cline{3-10}
  & &  B & M & Accuracy & Sensitivity / Recall & Specificity & AUC & Precision & F-score \\
  \hline \hline
A & Shen et al.,  2017 \cite{shen2015multi} (Multi-crop CNN)  & 528  & 297  & 87.14  & 77.00  & 93.00   &93.00  & Not given   & Not given \\
B & Hussein et al.,  2017 \cite{hussein2017risk} (3D CNN) & 635 &509  &91.26  &Not given  &Not given  &Not given  &Not given  &Not given\\
C & Han et al., 2015 \cite{han2015texture} (3D GLCM feature+SVM)  &1301 & 644 & 85.38$\pm$0.10 & 70.20$\pm$0.15 & 92.80$\pm$0.20 & 88.19$\pm$0.16 & 82.85$\pm$0.38 & 75.99$\pm$0.10\\
D & Dhara et al.,  2016 \cite{dhara2016combination} (Multi-visual features) & 1301 & 644 & 87.90$\pm$0.17 & 84.50$\pm$0.19 & 89.09$\pm$0.25 & 93.77$\pm$0.15  &79.31$\pm$0.37 & 81.82$\pm$0.21\\
E & Xie et al.,  2018 \cite{xie2018fusing} (Deep + visual features) & 1301 & 644  &88.73$\pm$0.15 & 84.40$\pm$0.20 & 90.88$\pm$0.13 & 94.02$\pm$0.20 & 82.09$\pm$0.24 & 83.23$\pm$0.21\\
F & Xie et al.,  2017 \cite{xie2017transferable} (TMME with Resnet-50) & 1301  &644 & 91.01$\pm$0.10  &83.83$\pm$0.15 & 94.56$\pm$0.13 & 95.35$\pm$0.15 & 88.40$\pm$0.24 & 86.07$\pm$0.15\\
G & Xie et al.,  2019a \cite{xie2018knowledge} (Knowledge-based)   & 1301 & 644  &91.60$\pm$0.15 & 86.52$\pm$0.25  &94.00$\pm$0.30  &95.70$\pm$0.24 & 87.75$\pm$0.52 & 87.13$\pm$0.16\\
H & Xie et al.,  2019b \cite{xie2019semi} (Semi-Supervised) & 1301  &644 & 92.53$\pm$0.05 & 84.94$\pm$0.17    & 96.28$\pm$0.08 & 95.81$\pm$ 0.19 & Not given & Not given  \\
I & Xu et al.,  2020 \cite{xu2020mscs}  (Multi-Scale Cost-Sensitive)  & 1156 & 556 & 92.64$\pm$ 0.12 & 85.58 $\pm$ 0.44 & 95.87$\pm$1.26 & 94.00 $\pm$0.26 & 90.39 $\pm$0.48 & 87.91 $\pm$0.11 \\
\midrule[1pt]
T1 & Low-Level-Feature \cite{ferrari2008learning}& 1301 & 644&88.73$\pm$0.15 & 84.40$\pm$0.20 & 90.88$\pm$0.13 & 94.02$\pm$0.20 & 82.09$\pm$0.24 & 83.23$\pm$0.21\\
T2 & Basic-visual-Feature \cite{kumar2009attribute} & 1301 & 644 & 91.01$\pm$0.10  &83.83$\pm$0.15 & 94.56$\pm$0.13 & 95.35$\pm$0.15 & 88.40$\pm$0.24 & 84.07$\pm$0.15\\
M1  & ResNet-50   & 1301 & 644 &88.14$\pm$0.23 & 82.17$\pm$0.14 & 90.77$\pm$0.15 & 91.21$\pm$0.14 & 82.18$\pm$0.11 & 82.36$\pm$0.14\\
M2  & Efficient-B4 & 1301 & 644 & 89.21$\pm$0.12   &83.83$\pm$0.24 & 91.18$\pm$0.22 & 92.05$\pm$0.27 & 86.88$\pm$0.19 & 83.04$\pm$0.23\\
M3  & ResNet-50-FPN   & 1301 & 644   &90.01$\pm$0.13 & 84.23$\pm$0.21 & 91.54$\pm$0.21 & 92.81$\pm$0.31 & 84.26$\pm$0.12 & 84.71$\pm$0.21\\
M4  & Efficient-B4-FPN   & 1301 & 644 & 90.91$\pm$0.22   &85.74$\pm$0.13 & 92.27$\pm$0.15 & 93.23$\pm$0.10 & 87.10$\pm$0.24 & 86.97$\pm$0.14\\
M31 & ResNet-50-FPN-GCN-Relation \cite{santoro2017simple} & 1301 & 644  &91.60$\pm$0.15 & 86.52$\pm$0.25  &92.32$\pm$0.15  &93.70$\pm$0.24 & 86.75$\pm$0.52 & 85.13$\pm$0.16\\
M41 & Efficient-B4-FPN-GCN-Relation \cite{santoro2017simple} & 1301  &644 & 92.15$\pm$0.12  &86.97$\pm$0.23 & 93.13$\pm$0.11 & 93.89$\pm$0.23 & 87.14$\pm$0.21 & 85.57$\pm$0.24\\
M32 & ResNet-50-FPN-GRU \cite{meng2018efficient} & 1301 & 644  &91.41$\pm$0.11 & 86.12$\pm$0.14  &92.92$\pm$0.19  &93.61$\pm$0.23 & 87.44$\pm$0.21 & 85.22$\pm$0.15\\
M42 & Efficient-B4-FPN-GRU  \cite{meng2018efficient} & 1301  &644 & 92.21$\pm$0.10  &86.94$\pm$0.11 & 93.89$\pm$0.24 & 93.72$\pm$0.12 & 88.24$\pm$0.13 & 86.63$\pm$0.21\\
O1 & Our-ResNet-50-FPN & 1301  &644 &93.74$\pm$0.17 & 89.23$\pm$0.21 & 95.76$\pm$0.24 & 96.12$\pm$0.12 & 94.21$\pm$0.14 & 88.27$\pm$0.31\\
O2 & Our-Efficient-B4-FPN & 1301  &644 & 95.31$\pm$0.15 & 90.51$\pm$0.15  & 96.15$\pm$0.22  & 96.47$\pm$0.31 & 95.95$\pm$0.24 & 88.83$\pm$0.45 \\
O2$^{*}$ & Our-Efficient-B4-FPN$^{*}$ & 1301  &644 & \textbf{95.36$\pm$0.10} & \textbf{91.01$\pm$0.16}  & \textbf{96.47$\pm$0.12}  & \textbf{96.54$\pm$0.32} & \textbf{95.96$\pm$0.21} & \textbf{89.13$\pm$0.15} \\
\bottomrule[2pt]
	\end{tabular}}
     \label{tab:lidcsota}
\end{table*}

\begin{table}[t]
    \caption{Comparison with Existing Lung Nodule Classification Models under a Second Training and Testing Protocol.}
	\centering
    \scalebox{0.9}{%
	\begin{tabular}{c|cc}
        \hline
	    Methods & Accuracy \% & \# of dataset \\
        \hline
        \hline
        TumorNet~\cite{hussein2017tumornet} & 82.47  & 1145  \\
        TumorNet-attribute~\cite{hussein2017tumornet} & 92.31 & 1145  \\
        SHC-DCNN~\cite{buty2016characterization}         & 82.40  & 1432  \\
        MCNN~\cite{shen2015multi} & 86.84 & 1100   \\
        CNN-MTL~\cite{hussein2017risk}   & 91.26  & 1340  \\
        PN-SAMP-S1~\cite{wu2018joint} & 92.03  & 1404 \\
        PN-SAMP-S2~\cite{wu2018joint} & 95.30  & 1404 \\
        PN-SAMP-M~\cite{wu2018joint}  & 97.58  & 1404  \\
        Ours &  \textbf{98.12}& 1404  \\
        \hline
        \hline
	\end{tabular}}
     \label{tab:lidcbotong}
\end{table}

\section{Medical Image Diagnosis}
In this section, we apply our attribute-based hybrid reasoning algorithm to two challenging medical image diagnosis tasks. The first task is benign-malignant classification of pulmonary nodules in chest computed tomography (CT) images. The second task is tuberculosis diagnosis using chest X-ray images. We define attributes and perform attribute relationship learning and reasoning for both tasks.

Following the common practice of previous attribute-based classification methods, not all attributes are used for training, i.e., we select attributes that have high causal relations with the diseases, and those unrelated attributes have been discarded.

\subsection{Pulmonary Nodule Classification}
Lung cancer gives rise to the most cancer-related deaths around the world~\cite{bray2018global}.
Early diagnosis and treatment are of great importance to long-term survival of lung cancer patients. Benign-malignant classification of pulmonary nodules in chest CTs is vital for timely diagnosis of lung cancer~\cite{ost2012decision}. In practice, radiologists often gain many clinical insights from the connections between domain knowledge and clinical symptoms as well as from the dependencies between diverse attributes and the underlying disease. This is applicable to pulmonary nodule diagnosis. For instance, when a nodule has an irregular shape and presents obvious spiculation, it has a high probability of being malignant.

\begin{table}[t]
\centering
\caption{Distribution of Median Malignancy Levels (MML) in the LIDC-IDRI Dataset for Lung Nodule Classification.}
\begin{tabular}{c|ccccc}
\toprule[1.5pt]
Dataset       & \multicolumn{2}{c}{Benign} & Uncertain & \multicolumn{2}{c}{Malignant}  \\
\midrule[1pt]
MML           & 1   & 2   & 3 & 4 & 5                       \\
\# of Nodule   & 358 & 943 & 612  & 474        & 170                         \\
\bottomrule[1.5pt]
\end{tabular}
\label{tab:lidcdataset}
\end{table}

\subsubsection{LIDC-IDRI Dataset} The LIDC-IDRI dataset~\cite{armato2011lung} from the Cancer Imaging Archive (TCIA) is one of the largest publicly available lung cancer datasets. It has 1018 clinical chest CT scans obtained from seven institutions. Each scan is associated with an XML file that details the locations of nodules on each 512$\times$512 slice. The diameters of the nodules range from 3mm to 30mm. Each suspicious lesion is categorized as a non-nodule, a nodule $<$ 3mm, or a nodule $\geq$ 3mm in diameter. Adopting the same setting as in \cite{han2015texture,dhara2016combination,xie2018knowledge,xie2019semi,xu2020mscs}, we only consider nodules $\geq$ 3mm in diameter since nodules $<$ 3mm are not considered to be clinically relevant by current screening protocols~\cite{han2015texture,dhara2016combination,hussein2017tumornet,shen2015multi,setio2016pulmonary,han2013texture}.

The malignancy of each nodule was evaluated with a 5-point scale, from benign to malignant, by up to four experienced thoracic radiologists. Following the same procedure used in previous studies~\cite{han2015texture,dhara2016combination,xie2018knowledge,xie2019semi,xu2020mscs}, we select those nodules annotated by at least one radiologist and calculate the median malignancy level (MML) for each of them. Nodules with an MML less than or higher than 3 are respectively labeled as benign or malignant. To reduce the uncertainty of nodule malignancy evaluation, nodules with an MML of 3 (noted as `uncertain') are excluded from our experiments as previous studies did. Thus there are a total of 1301 benign, 612 uncertain and 644 malignant nodules. The distribution of nodules over their MML is shown in Table~\ref{tab:lidcdataset}. Apart from malignancy, eight attributes are also graded for each nodule, i.e., subtlety, internal structure, calcification, sphericity, margin, lobulation, spiculation, and radiographic solidity. The grade of each attribute is between 1 and 5 except for internal structure (1-4) and calcification (1-6). We scale the grades of all attributes to 1-5 to maintain consistency, where 1 means the least obvious and 5 means the most obvious. Fig.~\ref{fig:att-lidc} shows sample 2D slices with distinct attributes.

We also normalize all chest CTs to a unified voxel size of $1.0 \times 1.0 \times 1.0 mm^3$ using spline interpolation. In addition, following the same setting as in \cite{xie2018knowledge,xie2019semi}, we assume the location of a nodule is the mean of the annotated centers of the nodule by radiologists. For each nodule, we crop a $64 \times 64 \times 64$ patch centered on its location.

\subsubsection{Experimental Setup}
In all experiments, we perform 10-fold cross validation, and each model is independently trained and tested 5 times with randomly initialized weights on each fold. The overall performance of a model is assessed with several commonly used metrics, including the mean and standard deviation of accuracy, sensitivity/recall, specificity, precision with the cut-off value of 0.5, F-score and AUC (area under the receiver operator curve).

All models are trained for 160 epoches from scratch using PyTorch~\cite{paszke2019pytorch} on NVIDIA Titan X pascal GPUs while Adam~\cite{kingma2014adam} being the optimizer with the initial learning rate set to 1e-3, which is reduced by a factor of 10 after every 30 epochs. The weight decay is set to 1e-4. Both ResNet and EfficientNet are adopted as the backbone models. We have implemented their 3D versions for the LIDC dataset by revising the original 2D versions. The size of the input image patches is $64 \times 64 \times 64$, and the batch size is 6 on a single GPU. Separate models are trained with and without data augmentation, which includes standard operations including flipping, rotation, and random cropping.

\subsubsection{Comparison with the State of the Art}
We have compared our hybrid algorithm with several state-of-the-art models for pulmonary nodule classification on the LIDC-IDRI dataset and the results are shown in Table~\ref{tab:lidcsota}, where O2 and O2$^{*}$ represent our models trained without and with standard data augmentation. The proposed method achieves the best performance under every evaluation metric when data augmentation is applied while it is still ranked first under every evaluation metric except for specificity when data augmentation is not used. Such a performance demonstrates the effectiveness of hybrid neural and probabilistic reasoning proposed in this paper, as well as the important role of attribute learning in medical image diagnosis. Specifically, method A adopts a multi-crop convolutional neural network trained a subset of 825 nodules from LIDC and achieves an accuracy of 87.14\%. Method B trains a 3D CNN with a subset of 1144 nodules and achieves a higher accuracy (91.26\%). All methods from C to H train models using the same number of nodules, which is 1945. However, method H introduced additional 1839 unlabeled nodules for semi-supervised learning and achieves the highest accuracy (92.53\%) among methods from C to H. Method I proposes a multi-scale 3D ResNet with a cost-sensitive loss and delivers a higher performance (92.64\%) than method H using a subset of 1712 nodules. In comparison, when adopting the same backbone (3D ResNet) as method I, our method (O1) achieves a better performance (93.74\%), indicating the importance of attribute relationship modeling in enhancing the performance of pulmonary nodule classification. Most importantly, Table~\ref{tab:lidcsota} shows that when using an EfficientB4-FPN backbone, our model with data augmentation (O2*) achieves the highest accuracy of 95.36\%, and our model without data augmentation (O2) gains 4.41\% over the baseline using the same backbone (M4).

\begin{figure*}[t]
\begin{center}
\includegraphics[width=1\linewidth]{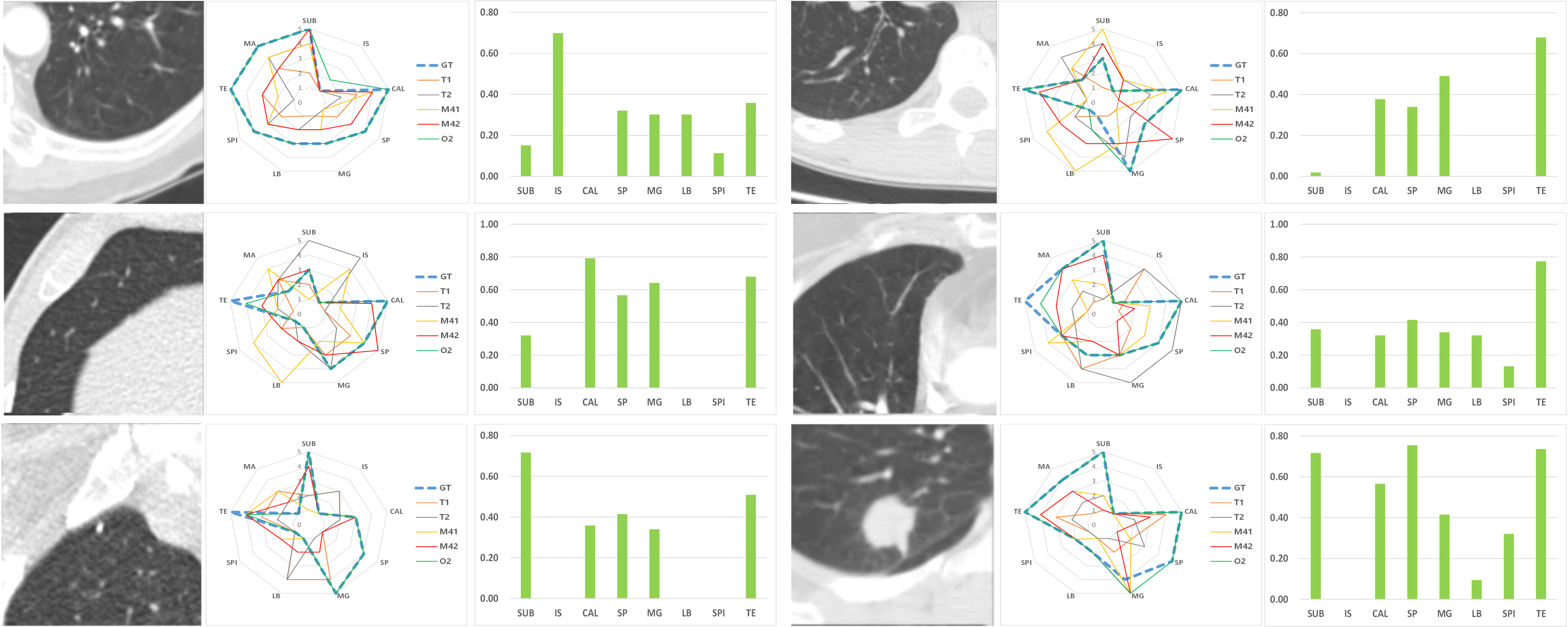}
\end{center}
\caption{Sample results of our method (O2) and four baseline methods (T1, T2, M41, M42) on the LIDC-IDRI dataset. Short names are defined as follows: SUB (Subtlety), IS (Internal Structure), CAL (Calcification), SP (Sphericity), MG (Margin), LB (Lobulation), SL (Spiculation), TE (Texture), and MA (Malignancy).
The radar charts visualize the actual grades of attributes and disease, and there exist five possible grades for each attribute or disease. GT stands for the ground truth.
The Histograms denote the numerical importance of different attributes for diagnosing malignancy (MA).
}\label{fig:dis-lidc}
\end{figure*}

\begin{figure}[t]
\begin{center}
\includegraphics[width=1\linewidth]{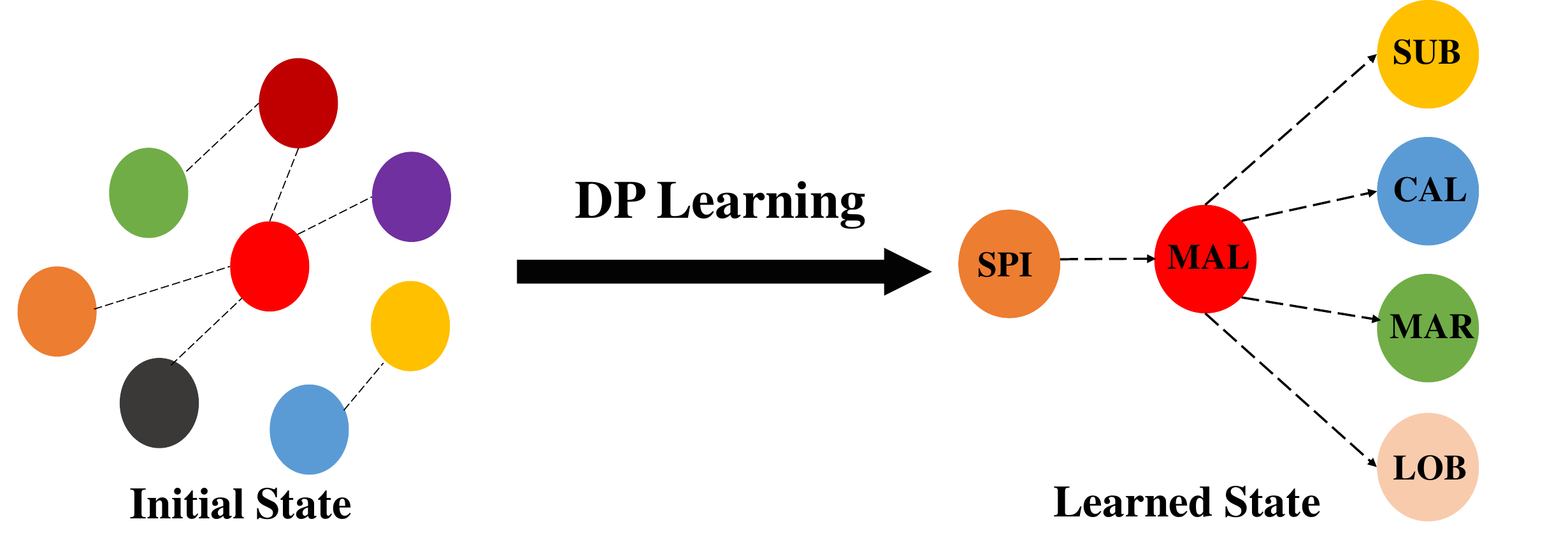}
\end{center}
\caption{The Bayesian network structure learned from the LIDC-IDRI dataset. Notations are defined as follows. SPI: spiculation; MAL: malignancy; SUB: subtlety; CAL: calcification; MAR: margin; LOB: lobulation.}\label{fig:bn}
\end{figure}

\begin{table*}[t]\normalsize
	\caption{Ablation Study of Our Classification Algorithm on the LIDC-IDRI Dataset. \textbf{`BN-1'} And \textbf{`BN-2'} Denote the First and Second BN Modules, \textbf{`CNA-RES'} Denotes Cross-Network Attention and Residual Fusion, \textbf{`SEatt'} Means Adopting SE Channel-Wise Attention in the GCN, \textbf{`GradBN'} Means Gradient Backpropagation through the BN Modules, \textbf{`AlterTrain'} Denotes the Proposed Alternating Training Strategy. \textbf{`Relation'} Means Removing Both BN Modules and Appending the Relational Network from \cite{santoro2017simple} after the GCN Module.}
	\centering
\scalebox{0.7}{%
\begin{tabular}{cccccccc|cccccc}
\toprule[2pt]
  GCN & BN-2 & CNA-RES & SEatt & BN-1 & GradBN & AlterTrain & Relation & Accuracy & Sensitivity / Recall & Specificity & AUC & Precision & F-score\\
\hline\hline
 $\checkmark$ &$\checkmark$ &  $\checkmark$  &  $\checkmark$  &  $\checkmark$ &  $\checkmark$  & $\checkmark$   &      $\boxtimes$  & \textbf{95.31$\pm$0.15} & \textbf{90.51$\pm$0.15}  & \textbf{96.15$\pm$0.22}  & \textbf{96.47$\pm$0.31} & \textbf{95.95$\pm$0.24} & \textbf{88.83$\pm$0.45} \\
 $\checkmark$ & $\boxtimes$            &  $\checkmark$  &  $\checkmark$  &  $\checkmark$  & $\checkmark$   &  $\checkmark$&     $\boxtimes$   & 94.74$\pm$0.09$^{\textbf{$\downarrow$}0.57}$  &88.92$\pm$0.13 & 95.86$\pm$0.41 & 95.83$\pm$0.09 & 92.54$\pm$0.06 & 88.04$\pm$0.04\\
 $\checkmark$&   $\boxtimes$    &    $\boxtimes$        & $\checkmark$   &    $\checkmark$            &  $\checkmark$               &$\checkmark$  &  $\boxtimes$  & 92.11$\pm$0.14$^{\textbf{$\downarrow$}3.20}$  & 88.14$\pm$0.22 & 93.74$\pm$0.15 & 94.91$\pm$0.26 & 88.11$\pm$0.09 & 86.88$\pm$0.12\\
$\boxtimes$& $\boxtimes$ &   $\boxtimes$ &  $\boxtimes$ &  $\checkmark$  & $\checkmark$   &  $\checkmark$  &     $\boxtimes$      & 91.21$\pm$0.22$^{\textbf{$\downarrow$}4.10}$   & 87.26$\pm$0.14 & 93.21$\pm$0.13 & 93.47$\pm$0.54 & 88.01$\pm$0.09 & 86.07$\pm$0.56\\
 $\checkmark$ &$\checkmark$ &       $\boxtimes$   &  $\checkmark$  &  $\checkmark$  & $\checkmark$   &  $\checkmark$&$\boxtimes$ & 93.64$\pm$0.11$^{\textbf{$\downarrow$}1.67}$  &88.01$\pm$0.15 & 95.03$\pm$0.13 & 95.02$\pm$0.13 & 91.42$\pm$0.13 & 87.92$\pm$0.04\\
 $\checkmark$& $\checkmark$ &  $\checkmark$  &    $\boxtimes$    &  $\checkmark$  & $\checkmark$   &  $\checkmark$& $\boxtimes$&94.01$\pm$0.02$^{\textbf{$\downarrow$}1.30}$  & 88.64$\pm$0.08 & 95.23$\pm$0.07 & 95.32$\pm$0.11 & 91.11$\pm$0.04 & 87.13$\pm$0.09\\
 $\checkmark$&$\checkmark$ &        $\boxtimes$    &  $\checkmark$  &  $\boxtimes$& $\checkmark$  &  $\checkmark$&   $\boxtimes$  &92.41$\pm$0.14$^{\textbf{$\downarrow$}2.90}$  & 88.14$\pm$0.22 & 93.74$\pm$0.15 & 94.91$\pm$0.26 & 92.11$\pm$0.09 & 86.88$\pm$0.12\\
  $\checkmark$&$\checkmark$ &  $\checkmark$  &  $\checkmark$  &  $\checkmark$  & $\boxtimes$      &  $\checkmark$&  $\boxtimes$   &93.81$\pm$0.10$^{\textbf{$\downarrow$}1.50}$  & 88.04$\pm$0.32 & 93.74$\pm$0.15 & 94.91$\pm$0.26 & 90.21$\pm$0.07 & 86.96$\pm$0.10\\
  $\checkmark$ & $\checkmark$ &  $\checkmark$  &  $\checkmark$  &  $\checkmark$  &    $\checkmark$            & $\boxtimes$ &$\boxtimes$    &94.31$\pm$0.10$^{\textbf{$\downarrow$}1.00}$  & 88.28$\pm$0.41 & 94.23$\pm$0.12 & 94.87$\pm$0.76 & 89.11$\pm$0.22 & 87.03$\pm$0.20\\
   $\checkmark$&    $\boxtimes$     & $\boxtimes$     & $\boxtimes$&$\boxtimes$&  $\boxtimes$    &$\boxtimes$&   $\checkmark$    & 92.15$\pm$0.12$^{\textbf{$\downarrow$}3.16}$  &86.97$\pm$0.23 & 93.13$\pm$0.11 & 93.89$\pm$0.23 & 87.14$\pm$0.21 & 85.57$\pm$0.24\\
    $\checkmark$&  $\boxtimes$ &$\boxtimes$& $\boxtimes$&$\boxtimes$&       $\boxtimes$   &$\boxtimes$& $\boxtimes$& 91.31$\pm$0.12$^{\textbf{$\downarrow$}4.00}$    &87.83$\pm$0.24 & 93.15$\pm$0.13 & 94.07$\pm$0.12 & 86.65$\pm$0.12 & 87.04$\pm$0.13\\
\bottomrule[2pt]
\end{tabular}}\label{tab:lidcabl}
\end{table*}

In addition, Table~\ref{tab:lidcbotong} shows that under the training setting and testing protocol adopted in \cite{wu2018joint}, our algorithm also achieves the highest accuracy, further demonstrating the effectiveness of our method. In this comparison, we adopt the same backbone (DenseNet) as in \cite{wu2018joint}, and do not apply any data augmentation during training. During testing, we adopt the same ``off-by-one'' accuracy as in \cite{wu2018joint}, which considers attribute/malignancy grading within $\pm$1 of the ground truth as correct results. Note that although we do not exploit the additional supervision signal provided by the nodule segmentation masks, that are used in \cite{wu2018joint}, our method still achieves the highest accuracy of 98.12\%.

We have further compared with attribute relationship modeling methods including M31, M41, M32, and M42. For a fair comparison, methods in the same comparison always adopt the same backbone, which is either 3D ResNet-50 or EfficientNet-B4. The relational network module proposed in \cite{santoro2017simple} and the GRU module proposed in \cite{meng2018efficient} respectively achieve an improvement of 1.59\% and 1.30\% in accuracy when compared with the 3D ResNet-50 baseline while our method improves the accuracy of the same baseline by 3.63\%. The same conclusion can be drawn for the EfficientNet-B4 backbone. We have also compared with two classic attribute learning methods T1 and T2, and they deliver a lower accuracy of 88.73\% and 91.01\%, respectively. In summary, all the above well-designed experiments demonstrate that the proposed hybrid neuro-probabilistic reasoning algorithm achieves clearly better performance than existing methods on the LIDC-IDRI dataset.

Sample input images and their associated attribute and disease classification results from our method and a few other baseline methods are shown in Fig.~\ref{fig:dis-lidc}. In addition, we provide the importance of individual attributes for pulmonary nodule classification, \emph{i.e.,} quantitatively evaluating how important individual attributes are in making the prediction. To be specific, we measure how much the final disease classification probability is affected by deactivating one attribute at a time in the input signals to BN-1 and GCN. For the input signals to GCN, an attribute can be deactivated by setting its corresponding feature vector to an average feature for that attribute over the samples where that attribute attains the lowest possible grade. For the input signals to BN-1, since each attribute in the LIDC-IDRI dataset is associated with 5 grades, deactivation means assigning the lowest grade (1) to the attribute.

\begin{figure*}[t]
\begin{center}
\includegraphics[width=1\linewidth]{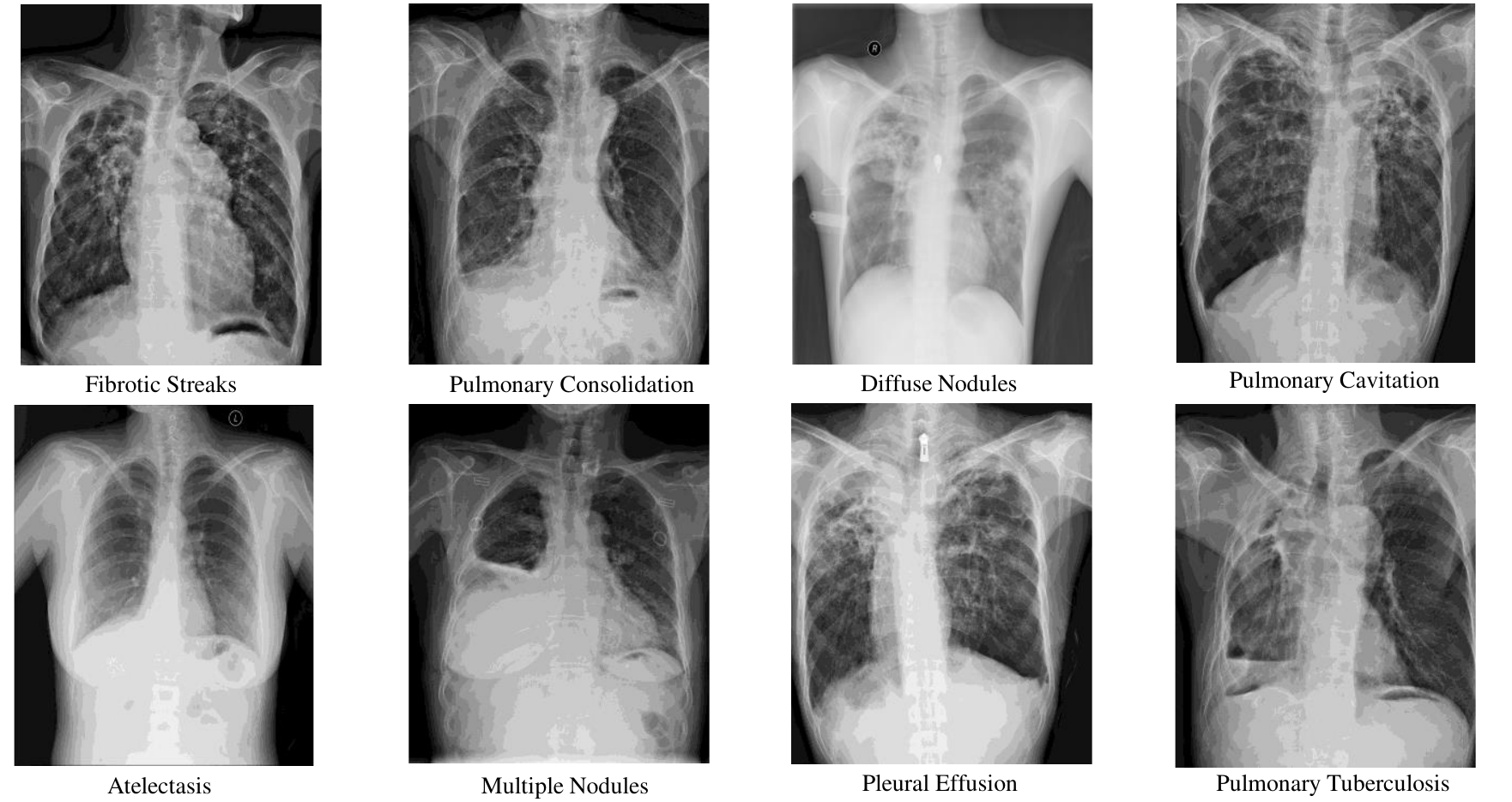}
\end{center}
\caption{Sample chest X-ray images used for building the TB-Xatt dataset, each with one of the following seven attributes except the last one: Fibrotic Streaks, Pulmonary Consolidation, Diffuse Nodules, Pulmonary Cavitation, Atelectasis, Multiple Nodules, and Pleural Effusion. The last image shows a case of the disease, Pulmonary Tuberculosis.}\label{fig:att-tb}
\end{figure*}

\subsubsection{Ablation Study} We have performed a systematic ablation study on the LIDC dataset to verify the effectiveness of the components in our framework. The results are shown in Table~\ref{tab:lidcabl}. Specifically, we have explored the effects of both the first and second BN (BN-1 and BN-2), cross-network attention and residual fusion (CNA-RES), SE channel-wise attention (SEatt)~\cite{hu2018squeeze}, gradient backpropagation through BN (GradBN), BN and DNN alternating training (AlterTrain) and the Relational Network Module (Relation)~\cite{santoro2017simple} on the EfficientNet-B4 backbone. Table~\ref{tab:lidcabl} shows that all proposed components or strategies contribute to the final performance of our algorithm. More specifically, we have found out that `CNA-RES' is the most important component, the performance would drop by 1.67\% if it was removed, which verifies the effectiveness of our proposed network coupling strategy. When both `BN-2' and `CNA-RES' are removed and the final result of the network is computed by a simple average of the results from BN-1 and GCN, the network becomes an ensemble model integrating the BN-1 and GCN branches. The performance drops significantly by 3.2\% under this ensemble model setting. This result indicates that various components in our hybrid framework are highly coupled, and altogether they deliver a performance much better than a simple ensemble model. `GradBN' and `AlterTrain' also have a significant impact, and the performance drops by 1.5\% and 1.0\% respectively if either of them is removed.
In addition, completely removing BN-1 and its coupling with GCN would decrease the performance by 2.90\% while removing BN-2 alone would decrease the performance by 0.57\%.
Moreover, the performance would drop by 3.16\% when both BN modules are removed and a relational network is appended after the GCN module. This result demonstrates the strong dependency modeling ability of BN modules. Fig.~\ref{fig:bn} shows the structure of the first BN module (BN-1) learned from the LIDC-IDRI dataset. This learned BN structure indicates that the node for malignancy has one parent node (i.e., spiculation) and four children nodes (i.e., subtlety, calcification, margin and lobulation).

We can draw the following conclusions from the results given in Table~\ref{tab:lidcabl},
\begin{itemize}
  \item `BN-1', `CNA-RES', `GradBN', `AlterTrain' and `SENet' are important to the classification performance on the LIDC-IDRI dataset. Note that removing `AlterTrain' actually means using ground-truth attribute labels instead of soft attribute labels for training the BN modules because the BN structures will not change if we always use ground-truth attribute labels during alternating training.
  \item BN modules are vital for attribute relational modeling. Their performance surpasses the performance of the advanced relational network.
  \item Although a simple ensemble model consisting of BN and GCN can already improve the performance, our proposed cross-network attention mechanism takes full advantage of BN and GCN, thus performs better.
\end{itemize}

\begin{table*}[t]
\centering
\caption{Distribution of Disease and Radiological Abnormalities in the TB-Xatt Dataset for Tuberculosis Diagnosis.}
\label{tab:chestdataset}
\scalebox{0.75}{%
\begin{tabular}{c|cccccccc}
\toprule[1.5pt]
Sign/Disease       & Fibrotic Streaks & Pulmonary Consolidation & Diffuse Nodules & Pulmonary Cavitation & Atelectasis & Multiple Nodules & Pleural Effusion & Pulmonary Tuberculosis   \\
\midrule[1pt]
\#Images        & 1,640 & 1,200  & 1,500  & 1,400    & 800  & 1,200        & 900        & 1,972               \\
\bottomrule[1.5pt]
\end{tabular}}
\end{table*}

\subsection{Tuberculosis Diagnosis}
Chest radiographs are a type of medical images that can be conveniently acquired for disease diagnosis. Radiologists can find out many diseases quickly via observing chest X-ray images, which are useful for early diagnosis and intervention. Tuberculosis~\cite{lin2014tuberculosis} has the highest mortality around the globe among infectious diseases. However, if it was detected at an early stage from chest radiographs, the death rate could decrease by 70\%~\cite{hart1977bcg}. Meanwhile, tuberculosis is a type of bacterial infection that can give rise to multiple types of radiological abnormalities in chest radiographs such as diffuse nodules and fibrotic appearance. Just like morphological characteristics of lung nodules, these radiological abnormalities can be treated as attributes to facilitate the diagnosis of tuberculosis by medical experts. Therefore, we choose tuberculosis diagnosis as an additional task to further evaluate the proposed hybrid neuro-probabilistic reasoning algorithm, which is able to model the dependencies between tuberculosis and radiological abnormalities to improve the reasoning capability of deep learning methods.

\begin{figure}[t]
\begin{center}
\includegraphics[width=1\linewidth]{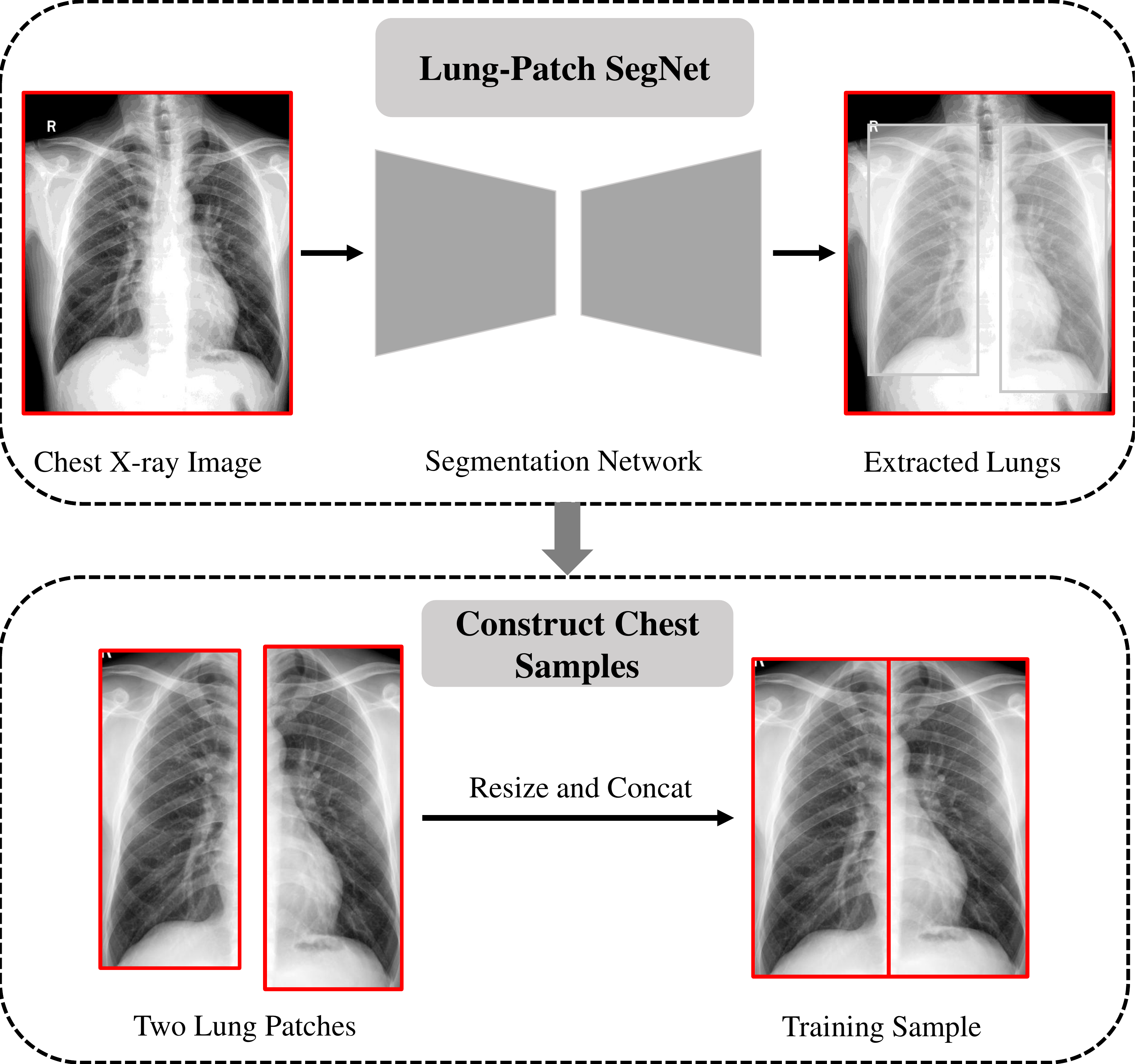}
\end{center}
\caption{Construction of image samples in the TB-Xatt dataset. First, lung segmentation in chest X-ray images using a trained segmentation model. Second, resizing the left and right lung patches to the same resolution and placing them side by side. }\label{fig:seglung}
\end{figure}

\begin{table*}[t]\normalsize
	\caption{Performance Comparison of Tuberculosis Diagnosis Models on the TB-Xatt Dataset.}
\centering
\scalebox{0.85}{%
	\centering
\begin{tabular}{c|c|c|c|c|c|c|c}
  \hline
\toprule[2pt]
  & \multirow{2}{*}{Methods}  & \multicolumn{6}{c}{Results(\%) (mean$\pm$standard deviation)} \\
 \cline{3-8}
  &  & Accuracy & Sensitivity / Recall & Specificity & AUC & Precision & F-score \\
  \hline \hline
A1 & Low-Level-Feature \cite{ferrari2008learning}&86.14$\pm$0.14 & 83.11$\pm$0.26 & 89.17$\pm$0.10 & 90.11$\pm$0.76 & 81.10$\pm$0.12 & 81.71$\pm$0.11\\
A2 & Basic-visual-Feature \cite{kumar2009attribute}  & 87.23$\pm$0.12  &84.16$\pm$0.27 & 92.00$\pm$0.09 & 91.27$\pm$0.63 & 83.26$\pm$0.71 & 83.44$\pm$0.17\\
S1 & Attention-Guide \cite{guan2018diagnose} & 93.12$\pm$0.10  &86.17$\pm$0.16 & 91.22$\pm$0.34 & 93.67$\pm$0.43 & 87.21$\pm$0.53 & 85.64$\pm$0.29\\
S2 & ADINet \cite{meng2020adinet} & 93.43$\pm$0.19  &87.24$\pm$0.25 & 91.87$\pm$0.11 & 94.11$\pm$0.29 & 87.44$\pm$0.54 & 84.21$\pm$0.13\\
\midrule[2pt]
M1 & ResNet-50 & 90.17$\pm$0.11 & 90.44$\pm$0.12 & 89.16$\pm$0.23 & 90.03$\pm$0.21 & 82.09$\pm$0.12 & 84.15$\pm$0.23\\
M2 & Efficient-B4 & 93.21$\pm$0.25  &91.54$\pm$0.22 & 92.55$\pm$0.11 & 92.67$\pm$0.19 & 84.27$\pm$0.98 & 87.23$\pm$0.09\\
M3 & ResNet-50-FPN & 91.23$\pm$0.28 & 91.96$\pm$0.42 & 91.22$\pm$0.33 & 91.54$\pm$0.07 & 83.27$\pm$0.08 & 85.26$\pm$0.44\\
M4 & Efficient-B4-FPN & 94.19$\pm$0.19  &92.78$\pm$0.09 & 92.97$\pm$0.01 & 94.01$\pm$0.10 & 86.11$\pm$0.02 & 89.17$\pm$0.08\\
\midrule[2pt]
M31 & ResNet-50-FPN-GCN-Relation \cite{santoro2017simple} & 91.74$\pm$0.29 & 92.16$\pm$0.32 & 93.23$\pm$0.25 & 92.26$\pm$0.12 & 86.14$\pm$0.11 & 86.54$\pm$0.13\\
M41 & Efficient-B4-FPN-GCN-Relation \cite{santoro2017simple}  & 95.17$\pm$0.11  &92.29$\pm$0.18 & 94.01$\pm$0.12 & 94.83$\pm$0.21 & 86.77$\pm$0.14 & 90.21$\pm$0.19\\
M32 & ResNet-50-FPN-GRU \cite{meng2018efficient} & 91.66$\pm$0.13 & 92.06$\pm$0.34 & 94.26$\pm$0.26 & 93.13$\pm$0.12 & 86.01$\pm$0.21 & 87.11$\pm$0.98\\
M42 & Efficient-B4-FPN-GRU \cite{meng2018efficient}& 94.65$\pm$0.32  &92.94$\pm$0.18 & 95.19$\pm$0.13 & 94.88$\pm$0.23 & 86.98$\pm$0.12 & 91.03$\pm$0.24\\
\midrule[2pt]
O1 & Our-ResNet-50-FPN &94.67$\pm$0.11 & 94.22$\pm$0.33 & 96.25$\pm$0.23 & 97.22$\pm$0.43 & 90.29$\pm$0.22 & 90.01$\pm$0.10\\
O2 & Our-Efficient-B4-FPN & 97.13$\pm$0.19 &95.51$\pm$0.10  & 97.11$\pm$0.16  & 98.10$\pm$0.34 & 92.44$\pm$0.31 & 91.78$\pm$0.27 \\
O2$^{*}$ & Our-Efficient-B4-FPN$^{*}$ & \textbf{98.41$\pm$0.10} & \textbf{96.32$\pm$0.11}  & \textbf{97.91$\pm$0.21}  & \textbf{98.91$\pm$0.12} & \textbf{93.24$\pm$0.10} & \textbf{92.77$\pm$0.14} \\
\bottomrule[2pt]
	\end{tabular}}
     \label{table:chestsota}
\end{table*}

\subsubsection{TB-Xatt dataset}
As for the task of tuberculosis diagnosis, there are a total of 14200 frontal chest X-ray images in an in-house dataset named TB-Xatt.
Every image in this dataset has been carefully annotated by certificated and experienced radiologists. Image annotations include 7 types of radiological abnormalities, `Pulmonary Consolidation', `Pulmonary Cavitation', `Diffuse Nodules', `Fibrotic Streaks', `Atelectasis', `Multiple Nodules', `Pleural Effusion' and one disease, `Pulmonary Tuberculosis'. On average, there are 3 occurrences of disease or radiological abnormalities per image and 1326 annotated images for each type of disease or radiological abnormalities. The distribution of images over disease and radiological abnormalities is presented in Table~\ref{tab:chestdataset}. In this study, we aim to improve the classification performance of `Pulmonary Tuberculosis' with the help of the seven types of radiological abnormalities, which are treated as the attributes of a chest X-ray image in our hybrid algorithm since these seven types of radiological abnormalities are the main expressions of active tuberculosis in the lung. Fig.~\ref{fig:att-tb} shows sample images with these attributes and disease from the TB-Xatt dataset. Note that the image samples in the dataset are pre-processed versions of the original X-ray images. We first perform lung segmentation in the original X-ray images and extract bounding boxes of the left and right lungs as shown in Fig.~\ref{fig:seglung}, then resize these bounding boxes to 512 $\times$ 256 and place each pair of left and right bounding boxes side by side as a single image sample.
In all experiments, we perform 10-fold cross validation, and each model is independently trained and tested 5 times with randomly initialized weights on each fold, the same as we did on LIDC.
The performance of a model is also assessed with the same set of metrics as LIDC, including accuracy, sensitivity/recall, specificity, precision, F-score and AUC.

\subsubsection{Experimental Setup} Different from LIDC, the ResNet and EfficientNet backbones are pre-trained on ImageNet. Adam is also the optimizer with the initial learning rate set to 1e-3, and the batch size is 32. All models are trained 60 epochs and the learning rate is reduced by a factor of 10 after 20 and 40 epochs.

\begin{figure*}[t]
\begin{center}
\includegraphics[width=1\linewidth]{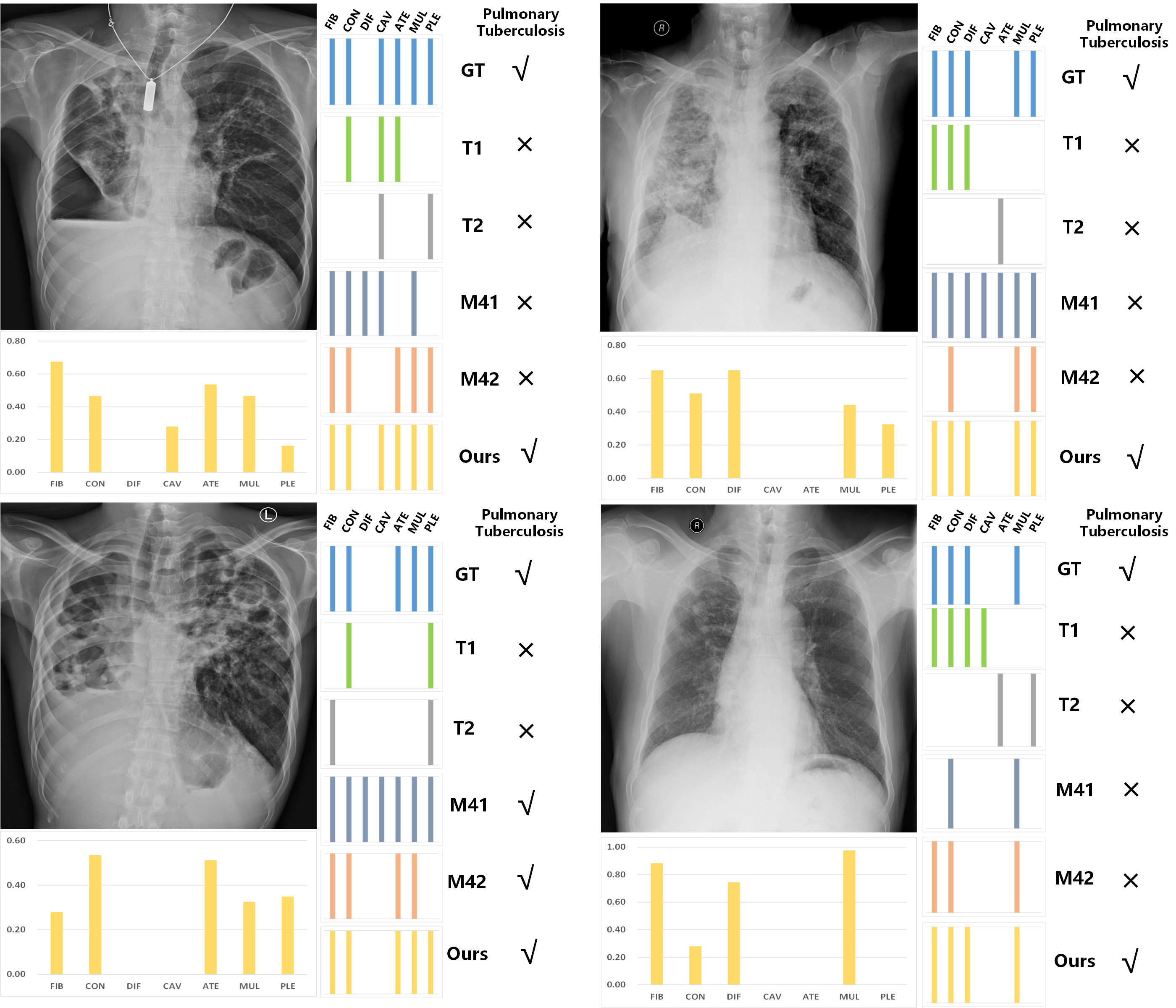}
\end{center}
\caption{Sample results of our method (O2) and four baseline methods (T1, T2, M41, M42) on the in-house TB-Xatt dataset. Short names are defined as follows: FIB (Fibrotic Streaks), CON (Pulmonary Consolidation), DIF (Diffuse Nodules), CAV (Pulmonary Cavitation), ATE (Atelectasis), MUL (Multiple Nodules), and PLE (Pleural Effusion).
The colored bars visualize attribute classification results from individual methods as well as the ground truth. GT stands for the ground truth.
$\checkmark$ and $\times$ visualize the diagnoses of pulmonary tuberculosis from individual methods as well as the ground truth.
The histograms under the X-ray images visualize the importance of individual attributes for diagnosing pulmonary tuberculosis.
}\label{fig:dis-tb}
\end{figure*}

\begin{table*}[t]\normalsize
	\caption{Ablation Study of Our Classification Algorithm on the TB-Xatt Dataset. \textbf{`BN-1'} And \textbf{`BN-2'} Denote the First and Second BN Modules, \textbf{`CNA-RES'} Denotes Cross-Network Attention and Residual Fusion, \textbf{`SEatt'} Means Adopting SE Channel-Wise Attention in the GCN, \textbf{`GradBn'} Means Gradient Backpropagation through the BN Modules, \textbf{`AlterTrain'} Denotes the Proposed Alternating Training Strategy. \textbf{`Relation'} Means Removing Both BN Modules and Appending the Relational Network from \cite{santoro2017simple} After the GCN Module.}
	\centering
\scalebox{0.7}{%
\begin{tabular}{cccccccc|cccccc}
\toprule[2pt]
  GCN & BN-2 & CNA-RES & SEatt & BN-1 & GradBN & AlterTrain & Relation & Accuracy & Sensitivity / Recall & Specificity & AUC & Precision & F-score\\
  \hline\hline
  $\checkmark$ & $\checkmark$ &  $\checkmark$  &  $\checkmark$  &  $\checkmark$  & $\checkmark$   &  $\checkmark$  &    $\boxtimes$        & \textbf{97.13$\pm$0.19} & \textbf{95.51$\pm$0.10}  & \textbf{97.11$\pm$0.16}  & \textbf{98.10$\pm$0.34} & \textbf{92.44$\pm$0.31} & \textbf{91.78$\pm$0.27} \\
  $\checkmark$& $\boxtimes$    &  $\checkmark$  &  $\checkmark$  &  $\checkmark$  & $\checkmark$   & $\checkmark$  &      $\boxtimes$       & 96.78$\pm$0.11$^{\textbf{$\downarrow$}0.35}$    & 94.26$\pm$0.13 & 96.78$\pm$0.23 & 97.12$\pm$0.11 & 90.93$\pm$0.17 & 90.03$\pm$0.07\\
  $\checkmark$& $\checkmark$ &   $\boxtimes$ &  $\checkmark$  &  $\checkmark$  & $\checkmark$   &  $\checkmark$  &     $\boxtimes$      & 96.21$\pm$0.44$^{\textbf{$\downarrow$}0.57}$   & 94.12$\pm$0.31 & 96.22$\pm$0.41 & 97.17$\pm$0.41 & 90.61$\pm$0.20 & 90.01$\pm$0.20\\
  $\boxtimes$& $\boxtimes$ &   $\boxtimes$ &  $\boxtimes$ &  $\checkmark$  & $\checkmark$   &  $\checkmark$  &     $\boxtimes$      & 95.69$\pm$0.12$^{\textbf{$\downarrow$}1.09}$   & 93.58$\pm$0.17 & 95.44$\pm$0.32 & 96.71$\pm$0.21 & 90.01$\pm$0.10 & 89.96$\pm$0.24\\
  $\checkmark$& $\checkmark$ &  $\checkmark$  &          $\boxtimes$           &  $\checkmark$  & $\checkmark$   &  $\checkmark$  &      $\boxtimes$      & 96.89$\pm$0.32$^{\textbf{$\downarrow$}0.24}$   & 94.47$\pm$0.55 & 96.79$\pm$0.42 & 97.87$\pm$0.31 & 91.02$\pm$0.33 & 90.64$\pm$0.25\\
  $\checkmark$& $\checkmark$ &  $\boxtimes$       &  $\checkmark$  &        $\boxtimes$             & $\checkmark$   &     $\checkmark$  &               $\boxtimes$      & 95.69$\pm$0.14$^{\textbf{$\downarrow$}1.44}$   & 94.55$\pm$0.43 & 94.78$\pm$0.11 & 95.94$\pm$0.12 & 87.11$\pm$0.23 & 91.21$\pm$0.54\\
  $\checkmark$& $\checkmark$ &  $\checkmark$  &  $\checkmark$  &  $\checkmark$  &           $\boxtimes$          &   $\checkmark$  &     $\boxtimes$      & 96.12$\pm$0.24$^{\textbf{$\downarrow$}0.66}$    & 94.66$\pm$0.31 & 94.51$\pm$0.23 & 95.17$\pm$0.21 & 90.14$\pm$0.32 & 90.11$\pm$0.23\\
  $\checkmark$& $\checkmark$ &  $\checkmark$  &  $\checkmark$  &  $\checkmark$  & $\checkmark$   & $\boxtimes$       &     $\boxtimes$       & 96.53$\pm$0.13$^{\textbf{$\downarrow$}0.60}$    & 94.71$\pm$0.24 & 94.62$\pm$0.33 & 95.19$\pm$0.47 & 90.28$\pm$0.34 & 90.51$\pm$0.33\\
  $\checkmark$&  $\boxtimes$    &           $\boxtimes$          &          $\boxtimes$           &         $\boxtimes$     &         $\boxtimes$          &    $\boxtimes$     &  $\checkmark$   & 94.97$\pm$0.21$^{\textbf{$\downarrow$}2.81}$  &93.09$\pm$0.08 & 93.46$\pm$0.29 & 94.93$\pm$0.21 & 86.77$\pm$0.14 & 90.21$\pm$0.19\\
\bottomrule[2pt]
\end{tabular}}
\label{tab:chestabl}
\end{table*}

\subsubsection{Comparison with the State of the Art}
On our in-house TB-Xatt dataset for tuberculosis diagnosis, we have also compared our hybrid algorithm with existing methods, including two state-of-the-art models (S1 and S2), two attribute learning methods (A1 and A2) and two relationship modeling methods (M31, M41 and M32, M42). The results are shown in Table~\ref{table:chestsota}, where O2 and O2$^{*}$ represent our models trained without and with standard data augmentation, respectively. No matter whether data augmentation is used, the proposed algorithm achieves the best performance under all evaluation metrics, including accuracy, sensitivity/recall, specificity, precision with the cut-off value of 0.5, F-score and AUC. Specifically, both A1 and A2 are classic attribute learning methods that use hand-crafted features, and achieve an accuracy of 86.14\% and 87.23\%, respectively. When compared with state-of-the-art disease classification models S1 and S2, the proposed hybrid algorithm demonstrates better performance with 4.01\% and 3.70\% improvement in accuracy, respectively. Furthermore, as in the pulmonary nodule classification task, we implemented two attribute relationship modeling methods (M31, M41 and M32, M42) based on the ResNet-50 and EfficientNet-B4 backbones. Results indicate that our method achieves better performance due to its strong causality modeling and relationship reasoning capabilities. The classification performance of our method with ResNet-50-FPN and EfficientNet-B4-FPN backbones are 94.67\% and 97.13\%, which respectively gain 3.44\% and 2.94\% over the two baselines using the same backbones. The results in Table~\ref{table:chestsota} demonstrate the effectiveness of our proposed algorithm on the tuberculosis diagnosis task. Sample input images and their associated attribute and disease classification results from our method and a few other baseline methods are shown in Fig.~\ref{fig:dis-tb}, where we have binary attribute and disease classifications. We also evaluate the importance of individual attributes for pulmonary tuberculosis diagnosis. Similarly, we measure how much the final disease classification probability is affected by deactivating one attribute at a time in the input signals to BN-1 and GCN. In this case, for the input signals to BN-1, deactivation means setting the classification probability of a specific attribute to zero. The deactivation scheme for the input signals to GCN is the same as that for the LIDC-IDRI dataset.

\begin{table*}[t]\normalsize
	\caption{Performance Comparison of Our Tuberculosis Diagnosis Models Trained with Increasingly Smaller Datasets. The Amount of Training Data in P1-P4 Has the Following Ratio, 1: 0.75 : 0.5: 0.25.}
\centering
\scalebox{0.88}{%
	\centering
\begin{tabular}{c|c|c|c|c|c|c|c}
  \hline
\toprule[2pt]
  & \multirow{2}{*}{Methods}  & \multicolumn{6}{c}{Results(\%) (mean$\pm$standard deviation)} \\
 \cline{3-8}
  &  & Accuracy & Sensitivity / Recall & Specificity & AUC & Precision & F score \\
  \hline \hline
\multirow{4}{*}{P1}  & Low-Level-Feature \cite{ferrari2008learning}&86.14$\pm$0.14 & 83.11$\pm$0.26 & 89.17$\pm$0.10 & 90.11$\pm$0.76 & 81.10$\pm$0.12 & 81.71$\pm$0.11\\ & Basic-visual-Feature \cite{kumar2009attribute}  & 87.23$\pm$0.12  &84.16$\pm$0.27 & 92.00$\pm$0.09 & 91.27$\pm$0.63 & 83.26$\pm$0.71 & 83.44$\pm$0.17\\
& Efficient-B4-FPN-GCN-Relation \cite{santoro2017simple}  & 94.97$\pm$0.21  &93.09$\pm$0.08 & 93.46$\pm$0.29 & 94.93$\pm$0.21 & 86.77$\pm$0.14 & 90.21$\pm$0.19\\
& Efficient-B4-FPN & 94.19$\pm$0.19  &92.78$\pm$0.09 & 92.97$\pm$0.01 & 94.01$\pm$0.10 & 86.11$\pm$0.02 & 89.17$\pm$0.08\\
 & Our-Efficient-B4-FPN & \textbf{97.13$\pm$0.19}$^{\textbf{$\uparrow$}2.94}$ & \textbf{95.51$\pm$0.10}  & \textbf{97.11$\pm$0.16}  & \textbf{98.10$\pm$0.34} & \textbf{92.44$\pm$0.31} & \textbf{91.78$\pm$0.27} \\
\midrule[1pt]
\multirow{4}{*}{P2} & Low-Level-Feature \cite{ferrari2008learning}&84.23$\pm$0.12 & 80.07$\pm$0.04 & 86.22$\pm$0.20 & 86.78$\pm$0.15 & 78.24$\pm$0.09 & 78.16$\pm$0.27\\
& Basic-visual-Feature \cite{kumar2009attribute}  & 83.45$\pm$0.22  &81.07$\pm$0.54 & 89.21$\pm$0.12 & 87.13$\pm$0.01 & 80.55$\pm$0.44 & 79.21$\pm$0.03\\
& Efficient-B4-FPN-GCN-Relation \cite{santoro2017simple}  & 91.26$\pm$0.34  &91.11$\pm$0.25 & 90.24$\pm$0.27 & 91.45$\pm$0.67 & 84.12$\pm$0.03 & 87.64$\pm$0.11\\
 & Efficient-B4-FPN & 91.21$\pm$0.14  &90.16$\pm$0.22 & 89.98$\pm$0.34 & 92.12$\pm$0.22 & 82.56$\pm$0.13 & 86.26$\pm$0.17\\
 & Our-Efficient-B4-FPN &93.21$\pm$0.44$^{\textbf{$\uparrow$}2.11}$ & 91.02$\pm$0.33  &91.20$\pm$0.21  &93.56$\pm$0.17 & 87.65$\pm$0.12 & 89.43$\pm$0.14\\
\midrule[1pt]
\multirow{4}{*}{P3} & Low-Level-Feature \cite{ferrari2008learning}& 80.11$\pm$0.09 & 75.67$\pm$0.12 & 80.32$\pm$0.55 & 81.26$\pm$0.64 & 72.13$\pm$0.07 & 72.51$\pm$0.62\\
& Basic-visual-Feature \cite{kumar2009attribute}  & 79.14$\pm$0.24  &78.27$\pm$0.63 & 82.11$\pm$0.10 & 80.65$\pm$0.44 & 76.17$\pm$0.23 & 74.56$\pm$0.35\\
& Efficient-B4-FPN-GCN-Relation \cite{santoro2017simple}  & 87.24$\pm$0.12  &86.10$\pm$0.65 & 87.21$\pm$0.46 & 87.13$\pm$0.55 & 81.21$\pm$0.46 & 83.27$\pm$0.35\\
 & Efficient-B4-FPN & 86.25$\pm$0.22  &84.23$\pm$0.27 & 84.22$\pm$0.54 & 82.45$\pm$0.32 & 79.15$\pm$0.21 & 80.24$\pm$0.55\\
 & Our-Efficient-B4-FPN &92.44$\pm$0.65$^{\textbf{$\uparrow$}6.19}$ & 90.01$\pm$0.21  &89.26$\pm$0.17  &89.22$\pm$0.10 & 86.14$\pm$0.25 & 85.23$\pm$0.17\\
\midrule[1pt]
\multirow{4}{*}{P4}& Low-Level-Feature \cite{ferrari2008learning}& 76.21$\pm$0.09 & 79.27$\pm$0.31 & 74.32$\pm$0.45 & 77.26$\pm$0.13 & 66.45$\pm$0.16 & 67.84$\pm$0.17\\
& Basic-visual-Feature \cite{kumar2009attribute}  & 75.67$\pm$0.54  &77.15$\pm$0.13 &76.55$\pm$0.43 & 76.21$\pm$0.22 & 72.56$\pm$0.34 & 71.27$\pm$0.21\\
& Efficient-B4-FPN-GCN-Relation \cite{santoro2017simple}  & 80.11$\pm$0.01  &78.20$\pm$0.15 & 81.31$\pm$0.11 & 80.13$\pm$0.24 & 75.56$\pm$0.31 & 76.27$\pm$0.45\\
 & Efficient-B4-FPN & 78.11$\pm$0.12  &79.26$\pm$0.36 & 78.23$\pm$0.34 & 80.15$\pm$0.24 & 76.27$\pm$0.15 & 73.62$\pm$0.14\\
 & Our-Efficient-B-FPN &86.98$\pm$0.27$^{\textbf{$\uparrow$8.87}}$ & 87.23$\pm$0.19  &87.63$\pm$0.24  &88.11$\pm$0.12 & 85.23$\pm$0.11 & 80.16$\pm$0.47\\
\bottomrule[2pt]
	\end{tabular}}
     \label{table:chestunlimited}
\end{table*}

\begin{table}[t]
\caption{Specification of Datasets Used for the Performance Comparison in Table~\ref{table:chestunlimited}.}
\centering
\label{tab:chestfolds}
\begin{tabular}{c|cccc}
\toprule[1.5pt]
Dataset       & P1 & P2 & P3 & P4  \\
\midrule[1pt]
\#Training Samples        & 11360 & 8530 & 5680 & 2840    \\
\#Validation Samples        & 1420   & 1420 & 1420 & 1420   \\
\#Testing Samples            & 1420   & 1420 & 1420 & 1420    \\
\midrule[1pt]
P/N ratio in Train     & 1/8 & 1/8    & 1/8  & 1/8    \\
P/N ratio in Val       & 1/3  &  1/3  &  1/3 &  1/3    \\
P/N ratio in Test      & 1/3  &  1/3  &  1/3 &  1/3    \\
\bottomrule[1.5pt]
\end{tabular}
\end{table}

\begin{center}
	\begin{table}[thb]
		\caption{Classification accuracy of input and output signals of BN-1 and GCN on TB-Xatt dataset.}\label{table:comment7}
		\centering
		\begin{tabular}{c|cc}
			\toprule[1.5pt]
			Method & P1  & P4 \\
			\hline
			BN-1 input & 91.04 & 78.22 \\
			BN-1 output & 93.10 & 85.44 \\
			GCN input & 92.12 & 78.23 \\
			GCN output & 96.36 & 80.28 \\
			Residual Fusion (GCN and BN-1) & 97.07 & 86.10 \\
			\bottomrule[1.5pt]
		\end{tabular}
	\end{table}
\end{center}

\subsubsection{Ablation Study} We have also conducted an ablation study on the TB-Xatt dataset to verify the effectiveness of the components in our proposed framework. The results are shown in Table~\ref{tab:chestabl}. As in the pulmonary nodule classification task, EfficientNet-B4 is the chosen backbone, and performance is measured when individual components are removed or replaced. Similar patterns in performance have been found here as well, and all the considered components contribute to the final performance of our hybrid algorithm.

Moreover, we have also tested the performance of our algorithm under the condition of a limited training set by sampling an increasingly smaller subset of the TB-Xatt dataset as the training set while keeping the size of the validation and testing sets unchanged. The sampling results are shown in Table~\ref{tab:chestfolds}, where P1 means all original training samples, and P2, P3 and P4 mean 75\%, 50\% and 25\% of the original training samples, respectively. For a fair comparison, we perform stratified sampling and ensure the same ratio between the number of positive and negative samples under all settings. This ratio is always 1/8 for the training sets, and 1/3 for the validation and testing sets. All results are shown in Table \ref{table:chestunlimited}, which indicates that the proposed algorithm can still perform better than the baselines when given a decreasing number of training samples. Note that all baselines adopt EfficientNet-B4 as the backbone since we empirically find out it is the best backbone under different settings. Furthermore, the performance gap between our algorithm and its corresponding baseline widens when the number of training samples decreases. When only 25\% of the original training samples are used, this performance gap reaches 8.87\% for the EfficientNet-B4 backbone. More surprisingly, although the number of training samples used in P4 is only half of that in P3, the performance of our models trained in the P4 setting is comparable and sometimes even better than the performance of corresponding baseline models in the P3 setting. These results demonstrate the strong generalization ability of our models trained using a small dataset.

We further investigate the classification performance of the individual output signals from the BN-1 and GCN modules to show how the two mechanisms work collaboratively when the size of training data varies.
As shown in Tab.~\ref{table:comment7}, (1) when the training dataset is small, BN-1 achieves better performance than GCN; (2) when the training dataset is large, GCN achieves better performance than BN-1. The justification is that (1) when the training dataset is small, BN-1 probabilistically models the causal relationships between attributes and diseases to mitigate the problem of insufficient data; (2) when the training dataset is large, GCN has the capability to learn expressive and powerful feature representations from the large-scale annotated dataset. Moreover, we notice that our residual fusion scheme always achieves higher accuracy than both individual networks (BN-1 and GCN).
It has become clear that the strong generalization ability of our trained models is made possible by the complementarity between our neural and probabilistic reasoning algorithms. The neural branch in our network realizes its full power when the training set is large while the probabilistic branch prevents overfitting from happening when the training set is small as probabilistic models are better at learning from smaller datasets.

\section{Conclusions and Discussions}
In this paper, we have introduced a hybrid neuro-probabilistic reasoning algorithm for verifiable attribute-based medical image diagnosis. There are two parallel branches in our hybrid algorithm, a Bayesian network branch performing probabilistic causal relationship reasoning and a graph convolutional network branch performing more generic relational modeling and reasoning using a feature representation. Tight coupling between these two branches is achieved via a cross-network attention mechanism and residual fusion of classification results. We train the hybrid network by alternatively updating neural network weights and the structure/parameters of the Bayesian networks.
We have successfully applied our hybrid reasoning algorithm to two challenging medical image diagnosis tasks, benign-malignant classification of pulmonary nodules in chest computed tomography images and tuberculosis diagnosis using chest X-ray images.  On the LIDC-IDRI benchmark dataset for pulmonary nodule classification, our method achieves a new state-of-the-art accuracy of 95.36\% and an AUC of 96.54\%. Our method also achieves a 2.94\% accuracy improvement on the in-house tuberculosis diagnosis dataset. Our ablation study indicates that our hybrid algorithm achieves a much more robust performance than a pure neural network architecture under very limited training data.

\appendices
\section{Gradient of Belief Propagation}\label{bpgradient}
Belief propagation is an iterative algorithm, where nodes exchange messages to update their marginal distributions until convergence. For a predefined Bayesian network without any loops, there exists a maximum number of iterations that guarantees convergence. The messages exchanged during belief propagation are computed using sums of products, which only involve additions and multiplications. Therefore, the entire belief propagation process is differentiable.

To be able to propagate gradients through Bayesian networks, we need to compute the gradient of all converged marginal distributions with respect to the network inputs, which are evidences at a subset of the nodes. Since both network inputs and outputs are multi-dimensional, the gradient we wish to compute is a Jacobian matrix. In the rest of this section, we outline an approach to compute this gradient. As there exist multiple variants of the belief propagation algorithm, without loss of generality, we formulate the gradient for the original belief propagation algorithm proposed by Pearl~\cite{pearl1982reverend,pearl2014probabilistic} for polytrees.

Let $x$ be the variable at a typical node $X$, which has $M$ children nodes, $\{Y_{1},..., Y_{M}\}$, and $N$ parents, $\{U_{1},..., U_{N}\}$. The current marginal distribution at $X$ is denoted as $P_B(x)$. There are two types of incoming messages at $X$. The first type, $\lambda_{Y_j}(x)$, is received from one of its children $Y_j$. It represents the current strength of the diagnostic support contributed by the outgoing link $X\rightarrow Y_j$. And the second type, $\pi_{X}(u_i)$, is received from one of its parents $U_i$. It represents the current strength of the causal support contributed by the incoming link $U_i\rightarrow X$. $P_B(x)$, $\lambda_{Y_j}(x)$ and $\pi_{X}(u_i)$ are all vectors of probabilities, representing discrete marginal or conditional distributions. Note that the evidence at a node can be modeled as an incoming message from an auxiliary child node. To facilitate gradient formulation, we synchronize all messages using time steps and every message has a time stamp. At each time step, every node updates its marginal distribution and outgoing messages as follows.
\begin{equation}
P_B^{t}(x) = \alpha \prod_{j}\lambda^{t-1}_{Y_j}(x) \sum_{u_{1},...,u_{n}}P(x|u_{1},...,u_{n})\prod_{i}\pi^{t-1}_{X}(u_i)
\label{eq:marginal}
\end{equation}
\begin{equation}
\begin{aligned}
\lambda^{t}_{X}(u_i) =& \beta \left[\sum_{x}\prod_{j}\lambda^{t-1}_{Y_j}(x)\right]  \\ & \ast \sum_{u_{k}:k\neq i}P(x|u_{1},...,u_{n})\prod_{k\neq i}\pi^{t-1}_{X}(u_{k})
\end{aligned}
\label{eq:bottom-up}
\end{equation}
\begin{equation}
\pi^t_{Y_{j}}(x) = \alpha \prod_{k\neq j}\lambda^{t-1}_{Y_{k}}(x) \sum_{u_1,...,u_{n}}P(x|u_{1},...,u_{n})\prod_{i}\pi^{t-1}_{X}(u_{i})
\label{eq:top-down}
\end{equation}
where $\alpha$ and $\beta$ are normalizing constants, $\lambda^{t}_{X}(u_i)$ is an outgoing message from $X$ to one of its parents, $\pi^t_{Y_{j}}(x)$ is an outgoing message from $X$ to one of its children, and $P(x|u_{1},...,u_{n})$ is the predetermined conditional probability table at $X$.

The marginal distribution and outgoing messages are defined as the variables associated with a node. Suppose the Bayesian network has $N_B$ nodes and $N_E$ edges, and each marginal or conditional distribution is defined on $L$ discrete grades. Since a polytree with $N_B$ nodes has $N_B-1$ edges and each edge is associated with two messages propagated along opposite directions, the total number of variables during belief propagation is $(N_B +2N_E)L=(3N_B-2)L$. According to (\ref{eq:marginal})-(\ref{eq:top-down}), at node $X$, the marginal distribution and outgoing messages at time step $t$ are all functions of the messages received from its parents and children at the previous time step. Thus, we can define a $(3N_B-2)L\times (3N_B-2)L$ Jacobian matrix, $J_t$, between every two consecutive time steps.
Note that $J_t$ is a sparse matrix. Suppose it takes $T$ time steps for belief propagation to reach convergence. The Jacobian matrix for all the time steps together, $J_{all}$, is thus a product of the Jacobian matrices for individual time steps, $J_{all}=\prod_{t=1}^{T} J_t$.
The gradient of the entire belief propagation algorithm, which consists of the partial derivatives of the final marginal distributions with respect to the input evidences, forms a submatrix of $J_{all}$, and can be extracted from $J_{all}$.

\bibliographystyle{splncs04}
\bibliography{ggsfnbib}

\end{document}